\documentclass[letterpaper, 10 pt, conference]{ieeeconf}  % Comment this line out if you need a4paper

\IEEEoverridecommandlockouts                              % This command is only needed if 
\usepackage[pscoord]{eso-pic}

\usepackage{longtable,tabularx}
\usepackage{tcolorbox}
\usepackage{xcolor,colortbl}
\usepackage{diagbox}
\usepackage{multirow}
\usepackage{svg}
\usepackage{graphicx} % for pdf, bitmapped graphics files
\usepackage{subcaption}
\usepackage{epsfig} % for postscript graphics files
\usepackage{amsmath} % assumes amsmath package installed
\usepackage{amssymb}  % assumes amsmath package installed
\usepackage{color}
\usepackage{cite}
\usepackage{booktabs}       % professional-quality tables

\usepackage[ruled, vlined, linesnumbered]{algorithm2e}
\usepackage{changepage}    
\usepackage[export]{adjustbox}

\usepackage{tikz}

\usetikzlibrary{automata,positioning,shapes,arrows,shadows,shadows.blur}
\newcommand{\shadowString}{circular drop shadow}

\usepackage{amsthm}
\newtheorem{definition}{Definition}
\newtheorem{problem}{Problem}
\newtheorem{example}{Example}
\newtheorem{remark}{Remark}

\DeclareMathOperator*{\argmax}{arg\,max}
\DeclareMathOperator*{\argmin}{arg\,min}

\DeclareMathOperator*{\cPreMax}{cPreMax}
\DeclareMathOperator*{\cPreMin}{cPreMin}
\DeclareMathOperator*{\PreImage}{PreImage}

\DeclareMathOperator*{\Compose}{Compose}

\DeclareMathOperator{\Val}{Val}
\DeclareMathOperator{\reg}{reg}

\newcommand{\R}{\mathbb{R}}
\newcommand{\B}{\mathcal{B}}
\newcommand{\G}{\mathcal{G}}
\newcommand{\PA}{\mathcal{P}}

\newcommand{\Tau}{\mathrm{T}}
\renewcommand{\phi}{\varphi}
\newcommand{\play}{P}
\definecolor{Gray}{gray}{0.85}
\definecolor{LightCyan}{rgb}{0.88,1,1}

\newcolumntype{a}{>{\columncolor{Gray}}c}
\newcolumntype{b}{>{\columncolor{white}}c}

\newcommand{\placetextbox}[3]{% \placetextbox{<horizontal pos>}{<vertical pos>}{<stuff>}
  \setbox0=\hbox{#3}% Put <stuff> in a box
  \AddToShipoutPictureFG*{% Add <stuff> to current page foreground
    \put(\LenToUnit{#1\paperwidth},\LenToUnit{#2\paperheight}){\vtop{{\null}\makebox[0pt][c]{#3}}}%
  }%
}%

\newif\ifarxiv
\arxivtrue
\ifarxiv
\fi

\title{\LARGE \bf
Efficient Symbolic Approaches for Quantitative Reactive Synthesis \\with Finite Tasks}

\author{Karan Muvvala and Morteza Lahijanian% <-this % stops a space
\thanks{This work was supported in part by the University of Colorado Boulder and NASA COLDTech Program under grant \#80NSSC21K1031.}%
\thanks{Authors are with the Department of Aerospace Engineering Sciences at the University of Colorado Boulder, CO, USA
        {\tt\small \{\textit{firstname}.\textit{lastname}\}@colorado.edu}}%
}

\begin{document}

\ifarxiv
\placetextbox{0.5}{0.95}{To appear in the IEEE/RSJ International Conference on Intelligent Robots and Systems (IROS), October 2023.}
\fi

% \AddToShipoutPictureBG*{%
%   \AtPageUpperLeft{%
%     % \hspace{16.5cm}%
%     \raisebox{-1.5cm}{%
%       \makebox[0pt][r]{Submitted to the IEEE/RSJ International Conference on Intelligent Robots and Systems (IROS), 2023.}}}}

\maketitle

\thispagestyle{plain}
\pagestyle{plain}

\begin{abstract}

This work introduces efficient symbolic algorithms for quantitative reactive synthesis.  
We consider resource-constrained robotic manipulators that need to interact with a human to achieve a complex task expressed in linear temporal logic.  
Our framework generates reactive strategies that not only guarantee task completion but also seek cooperation with the human when possible.  
We model the interaction as a two-player game and consider regret-minimizing strategies to encourage cooperation.  We use symbolic representation of the game to enable scalability. 
% using boolean variables and functions with both binary and algebraic decision diagram encodings,
For synthesis, 
we first introduce value iteration algorithms 
for such games with min-max objectives. Then, we extend our method to the regret-minimizing objectives.  Our benchmarks reveal that our the
symbolic framework not only significantly improves computation time (up to an order of magnitude) but also can scale up to much larger instances of manipulation problems
with up to 2$\times$ number of objects and locations than the state of the art.
\end{abstract}

\section{Introduction}
\label{sec: intro}

Robots are experiencing a boom in their deployment in the real world. Application domains include warehouse, assembly, delivery, home assistive, and planetary exploration robots. As they transition from robot-centric workspaces to the unstructured environments, robots must be able to achieve their tasks through interactions with other agents while dealing with finite resources. This is a computationally challenging problem because planning a reaction for every possible action of other agents, known as the \textit{reactive synthesis} problem, in face of resource constraints is expensive.  
The computational challenge is exacerbated in robotic manipulators where tasks are often complex and robots have high degrees of freedom.
More specifically, when faced with humans, the problem becomes even more difficult because we expect robots to have meaningful interactions with us while still trying to complete their tasks. 
% without exceeding their resource budget.

Our approach is based on modeling the robot-human interaction with resource constraints as a quantitative two-player game.  
We use \textit{Linear Temporal Logic over finite trace} (LTLf) \cite{vardi2013ltlf} for precise specification of tasks that can be accomplished in finite time. Using the game-theoretic approach allows us to base our method within the reactive synthesis framework to generate correct-by-construction strategies that guarantee satisfaction of the task while enabling the robot to have meaningful interaction with the human.  Specifically, we build on the results of \cite{muvvala2022regret}, which show resource-minimizing strategies emit ``unfriendly'' interactions because the human is viewed as an adversarial agent, and hence, these strategies attempt to avoid the human by all means.  Instead, \textit{regret-minimizing} strategies, which evaluate the cost of actions in the hindsight relative to a resource budget, are ``friendly'' because they seek cooperation with the human up until resources are endangered. 
Fig.~\ref{fig: strs_illustration} illustrates robot behaviors under these strategies. 
% The robot expends more energy to operate away from the human (1a) and less energy where the human can manipulate the object (1b).  In Fig. \ref{fig: adv_beh}, the robot finishes the task away from the human irrespective of the human’s intention, while in (1.b), the robot operates near the human if the expected behavior of the human is to be cooperative.}
% \ml{This explanation is not enough... where is the human region? which objects are movable?  Should perhaps explain everything in the caption of the figure rather than here.  Could just refer to the figure here in the text.}

% \km{Consider the Scenario in Fig. \ref{fig: strs_illustration}. The robot spends more energy per action to finish the task away from the human (1.b) and less energy per action in the region where the human can manipulate the object (1.a). In Fig. 1a, using that approach, the robot builds the arch away from the human irrespective of the human’s intention. However, using the approach in (1.b), the robot finishes the task near the human (Fig. 1b) if the expected behavior of the human is to be cooperative.}
However, the challenge with regret-minimizing strategies is that they are finite-memory \cite{muvvala2022regret}, requiring the knowledge of history to evaluate regret, leading to a computational blow-up (both in time and space) on an already-difficult problem.

%%%%  For Arxiv we use eps files
\ifarxiv
\begin{figure}[t]
    % \centering
    \begin{subfigure}[t]{0.5\linewidth}
        \centering
        \includegraphics[width=0.98\linewidth, valign=t]{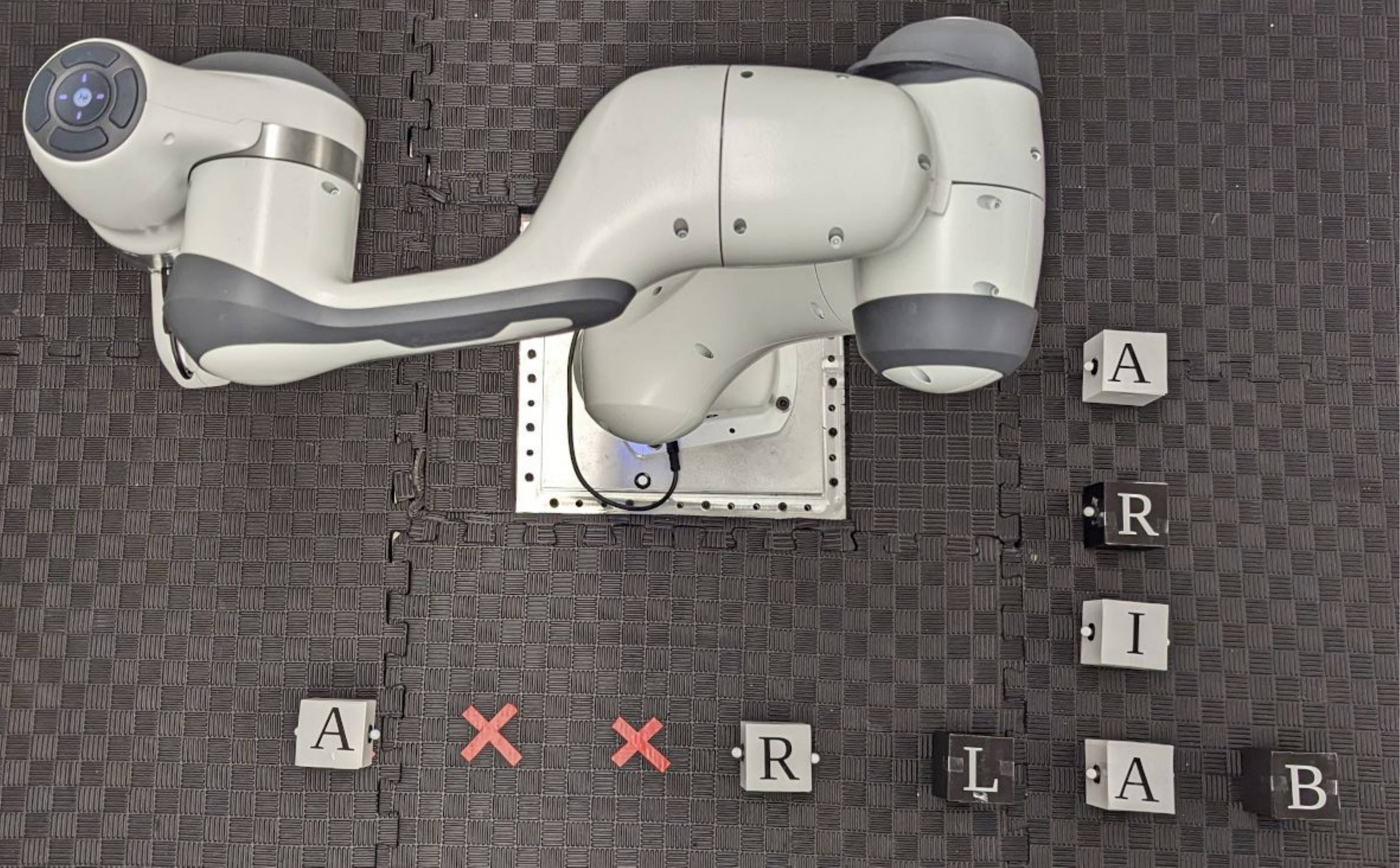}
        \caption{Adversarial Behavior}
        \label{fig: adv_beh}
    \end{subfigure}%
    \begin{subfigure}[t]{0.5\linewidth}
        \centering
        \includegraphics[width=0.98\linewidth, valign=t]{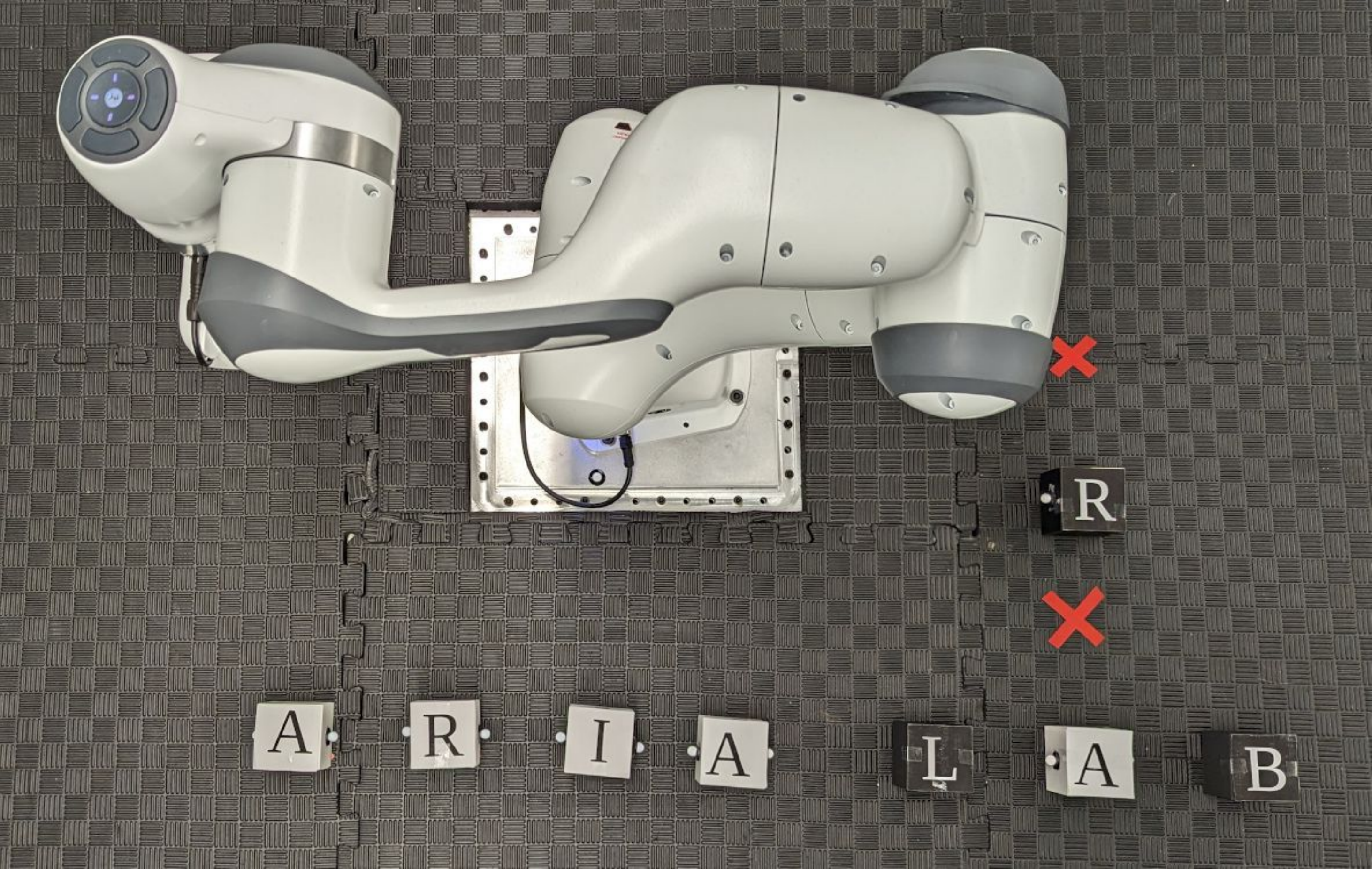}
        \caption{Regret Behavior}
        \label{fig: reg_beh}
    \end{subfigure}%
    \caption{The robot is tasked with spelling ``ARIA LAB" either horizontally or vertically. The white boxes are movable while the black boxes are fixed.  A human can reach and move boxes placed on the left of the ``L" box.
    (a) Resource-minimizing strategy via \cite{he2017reactive}, under which the robot expends more energy to operate away from the human.
    (b) Regret-minimizing strategy via \cite{muvvala2022regret}, under which the robot operates near the human to seek cooperation.}
    \label{fig: strs_illustration}
    % \vspace{-5mm}
\end{figure}

%%% Non Arxiv version uses jpg images.
\else
\begin{figure}[t]
    % \centering
    \begin{subfigure}[t]{0.5\linewidth}
        \centering
        \includegraphics[width=0.98\linewidth, valign=t]{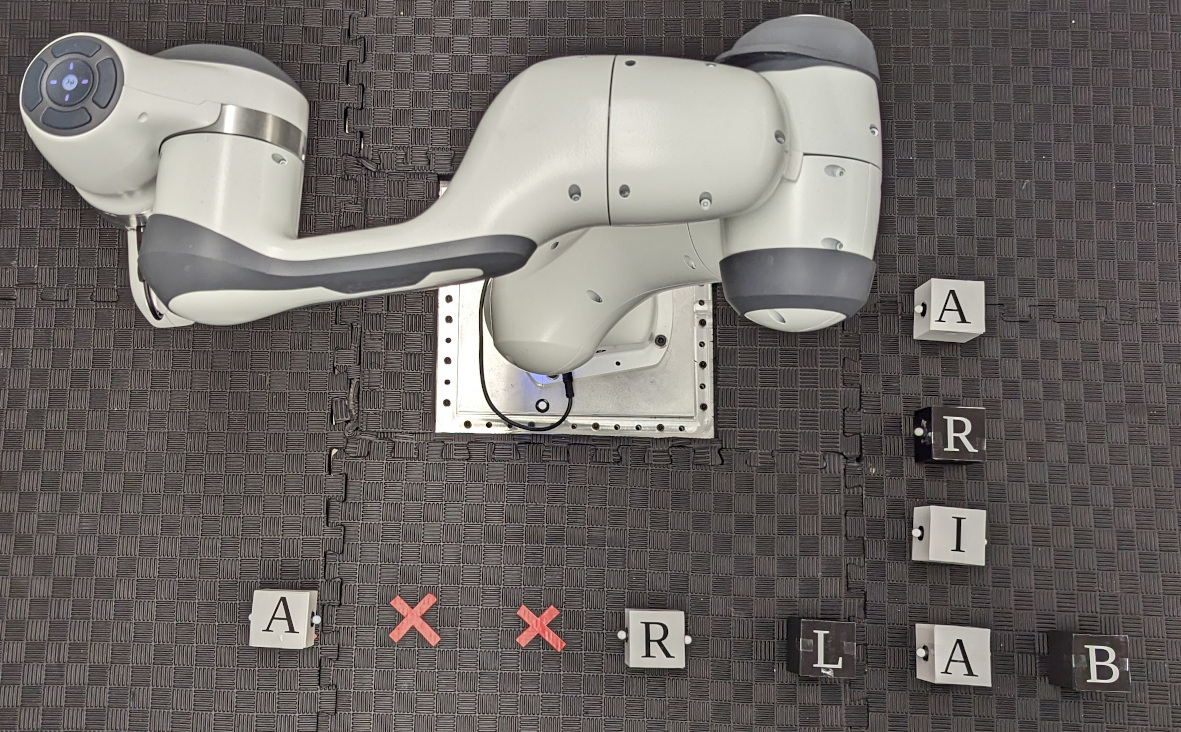}
        \caption{Adversarial Behavior}
        \label{fig: adv_beh}
    \end{subfigure}%
    \begin{subfigure}[t]{0.5\linewidth}
        \centering
        \includegraphics[width=0.98\linewidth, valign=t]{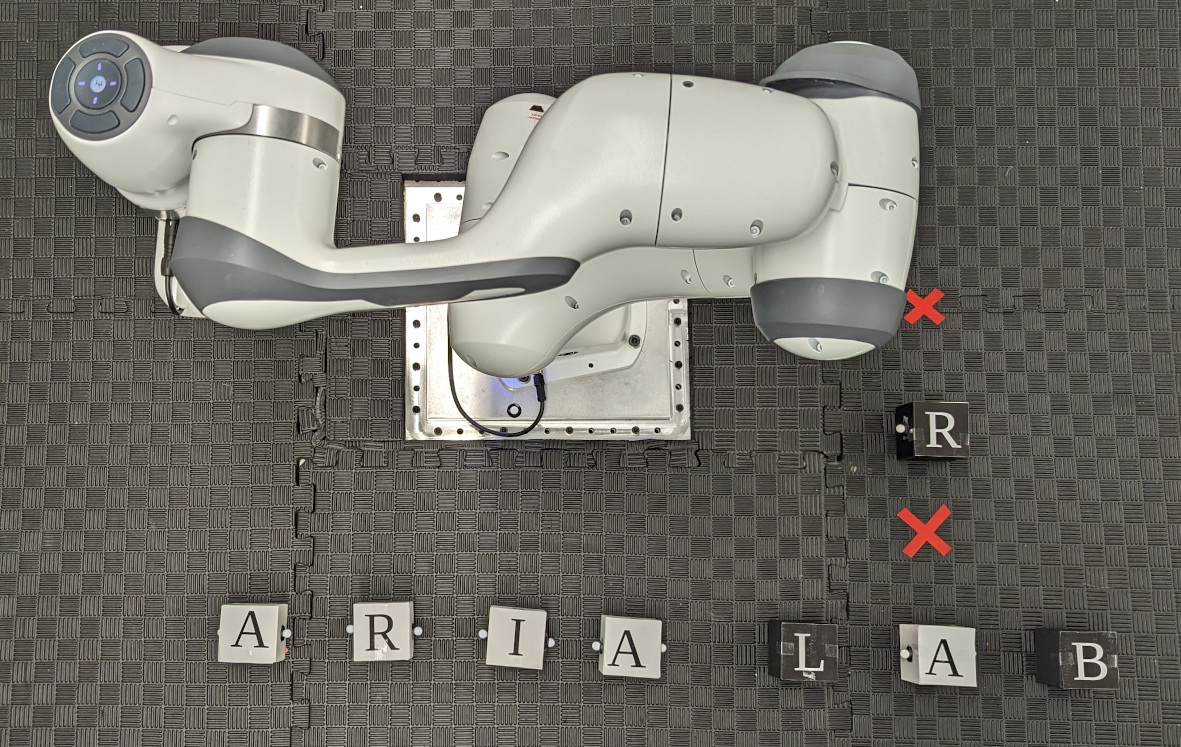}
        \caption{Regret Behavior}
        \label{fig: reg_beh}
    \end{subfigure}%
    \caption{
    % ``ARIA LAB" construction.
    The robot is tasked with spelling ``ARIA LAB" either horizontally or vertically. The white boxes are movable while the black boxes are fixed.  A human can reach and move boxes placed on the left of the ``L" box.
    (a) Resource-minimizing strategy via \cite{he2017reactive}, under which the robot expends more energy to operate away from the human.
    (b) Regret-minimizing strategy via \cite{muvvala2022regret}, under which the robot operates near the human to seek cooperation.}
    \label{fig: strs_illustration}
    % \vspace{-5mm}
\end{figure}
\fi

% \ml{fix this paragraph...}
To mitigate the computational bottleneck, we propose symbolic algorithms for quantitative games.  We represent the manipulation game symbolically, 
% using boolean variables and functions with both binary and algebraic decision diagram encodings,
which significantly reduces memory usage. Further, we introduce value iteration algorithms for quantitative symbolic games with min-max objectives. Finally, we extend our symbolic approach to the regret framework in \cite{muvvala2022regret}.  Case studies and benchmarks reveal that our symbolic framework not only significantly improves computation time (up to an order of magnitude) but also can scale up to way larger instances of manipulation problems with up to 
% 3$\times$ number of locations with
2$\times$ number of movable objects and locations
% \ml{4x larger number of objects of locations??}
than the state of the art.

In short, the contributions of this work are fourfold: 

(i) the \emph{first} symbolic value iteration algorithm using the compositional encoding for two-player quantitative games, to the best of our knowledge, 
(ii) an efficient symbolic  synthesis algorithm for regret-minimizing strategies,
(iii) an open-source tool for quantitative symbolic reactive synthesis with Planning Domain Definition Language (PDDL) plug-in, and
(iv) a series of benchmarks and illustrative manipulation examples that show the efficacy of the proposed symbolic framework in scaling up to real-world problems.

\subsection{Related Work}
Reactive synthesis has recently been studied for both mobile robots \cite{hadas2012reactive, kress2018synthesis} 
and manipulators \cite{he2015towards, hunter2017reactive}.
The approaches to mobile robots are based on qualitative games.  Work
\cite{he2019automated, he2017reactive} in robotic manipulation use quantitative turn-based game abstraction to model the interaction between the human and the robot. 
By assuming the human is adversarial, winning strategies are synthesized for the robot that guarantee task completion. While this approach has nice task completion guarantees with resource constraints, the adversarial assumption is conservative and results in one-sided, uncooperative interactions.
% and does not model the human as a strategic agent with its own objective, possibly not adversarial. 
Recently, \cite{wells2021probabilistic} relaxed this assumption by modeling the human as a probabilistic agent. They use data gathered from past interactions to model the interaction as a Markov decision process.  This approach results in better interactions, but the synthesized strategies reason about the human behavior in expectation, rather than as a strategic agent. In the most recent work \cite{muvvala2022regret}, the notion of \textit{regret} is employed to incorporate cooperation-seeking behaviors for the robot while still preserving the task completion guarantees. That approach produces ``meaningful'' interactions, but the robot needs to keep track of human's past actions, leading to major computational issues.  
% \ml{to save space for the final version, this paragraph can be summarized quite a bit}

To enable scalability of reactive synthesis, recent work \cite{he2019symbolic} focuses on symbolic approaches.  They use Binary Decision Diagrams (BDDs) to represent sets of states and transitions as boolean functions. With this representation, the robot can efficiently reason about the effects of its actions on the world by manipulating boolean formulas. Work \cite{he2019symbolic} shows how planning with LTLf \cite{zhu2017symbolic} symbolically allows much better scalability for qualitative manipulation games.  In \cite{chatterjee2014strategy}, a hybrid approach to symbolic min-max game is formulated. The states are represented explicitly, while their values are stored symbolically. Work \cite{maoz2016symbolic} extends this framework to purely symbolic algorithms for specifications defined over the infinite horizon GR(1). 
% However, such infinite tasks are not suitable for robotic manipulation tasks. 
% more computationally expensive to check and can be more difficult to use in practice.
% and hence are not suitable for the problem we are considering. 
% \ml{how is this different from ours other than the GR1 part?}
Although these approaches are effective, they are limited to infinite tasks. In manipulation, however, tasks are often finite (achievable in finite time), which is the focus of this work.
\section{Problem Formulation}
\label{sec: prob_form}

% In this work, our goal is to synthesize strategies for robotic manipulators to achieve a high-level task while interacting with a human in a shared workspace.   We specifically aim to \emph{efficiently} generate reactive strategies that not only guarantee task completion but also seek cooperation with the human when possible.

% Similar to previous approaches \cite{muvvala2022regret, he2017reactive}, we assume an abstraction to formally model the human and the robot interacting with each other. Finally, we use model-checking based 
% approaches to synthesize reactive strategies for the robot that guarantee task completion and generate cooperation-seeking behavior whenever possible. 

\subsection{Manipulation Domain Abstraction}

The problem of planning for a robot in interaction with a human is naturally continuous with high degrees of freedom. 
% To enable the use of tools developed by the model-checking community, we need to discretize the workspace
Due to difficulty of reasoning in such spaces, they are commonly abstracted to discrete models \cite{he2019automated,muvvala2022regret, he2017reactive, wells2021probabilistic}, known as abstraction.
Similar to \cite{muvvala2022regret, he2017reactive}, we use a two-player game abstraction to model the human-robot interaction. 
% \ml{space saving: this paragraph can be summarized in one sentence.}

\begin{definition}[Two-player Manipulation Domain Game]
    \label{def:game abstraction}
 The two-player turn-based manipulation domain is defined as a tuple  $\G = (V, v_0, A_s, A_e, \delta, F, \Pi, L)$, where
 \begin{itemize}
     \item $V = V_s \cup V_e$ is the set of finite states,  where $V_s$ and $V_e$ are the sets of the robot and human player states, respectively, such that $V_s \cap V_e = \emptyset$, 
     \item $v_0 \in V_s$ is the initial state,
     \item $A_s$ and $A_e$ are the sets of finite actions for the robot and human player, respectively,
     \item $\delta: V \times (A_s \cup A_e) \to V$ is the transition function
     such that,
     for $i,j \in \{s,e\}$ and $i\neq j$, given state $v\in V_i$ and action $a \in A_i$, the successor state is $\delta(v,a) \in V_j$,
     \item $F: V \times (A_s \cup A_e) \to \mathbb{R}_{\geq 0}$ is the cost function that, for robot $v\in V_s$ and $a\in A_s$, assigns cost $F(v,a) \geq 0$, and $F(v,a) = 0$ for all human $v\in V_e$ and $a\in A_e$,
     \item  $\Pi = \{p_0, \ldots, p_n\}$ is set of task-related propositions that can either be true or false, and
     \item $L: V \to 2^{\Pi}$ is the labeling function that maps each state $v \in V$ to a set of atomic propositions $L(v) \subseteq \Pi$ that are true in $v$.  
 \end{itemize}
\end{definition}

% We assume the abstraction to be non-blocking, i.e., from every state, there exists at least one action such that $\delta: V \times (A_s \cup A_e) \notin \varnothing$. 

\noindent
% \ml{need to relate the game with the manipulation domain...  fix this paragraph}
Intuitively, every state in $\G$ encodes the current configuration of the objects and the status of the robot. For instance, each state is a tuple that captures the locations of all the objects in the workspace and the end effector's status as free or occupied. A successor state encodes the status of the objects and the end effector under a human or robot action. 
% This is can naturall encode using preconditions  
% From a state under each human action is .... and robot action is .... .

\begin{example}
\label{ex: manip_domain}
Fig. \ref{fig: mani_ts} depicts a sample manipulation domain for a robot.  Fig. \ref{fig: mani_game} is its extension to a game abstraction $\G$ that captures the interaction between the robot and a human. 
% The circle and square states belong to the robot and human player, respectively. At each state, the respective player takes an action, and the game evolves in turns.
\end{example}

% \begin{figure}[t]
%     \centering
%     \includesvg[width=\linewidth]{images/manip_domain_v2.svg}
%     \caption{Manipulation Domain without human action}
%     \label{fig:my_label}
% \end{figure}

% \begin{figure}[t]
%     \centering
%     \includesvg[width=\linewidth]{images/manip_domain_w_human_v1.svg}
%     \caption{Manipulation Domain with human action}
%     \label{fig:my_label}
% \end{figure}

%%% For Arxiv version we use the PDF 1.4 version

\ifarxiv
\begin{figure}[t!]
    \centering
    \begin{subfigure}[t]{0.45\columnwidth}
        \centering
        %% This is the updated one but for some reaons is not being rendered correctly
        % \includegraphics[width=0.99\linewidth]{images/updated_manip_domain_v3.pdf}
        \includegraphics[width=0.99\linewidth]{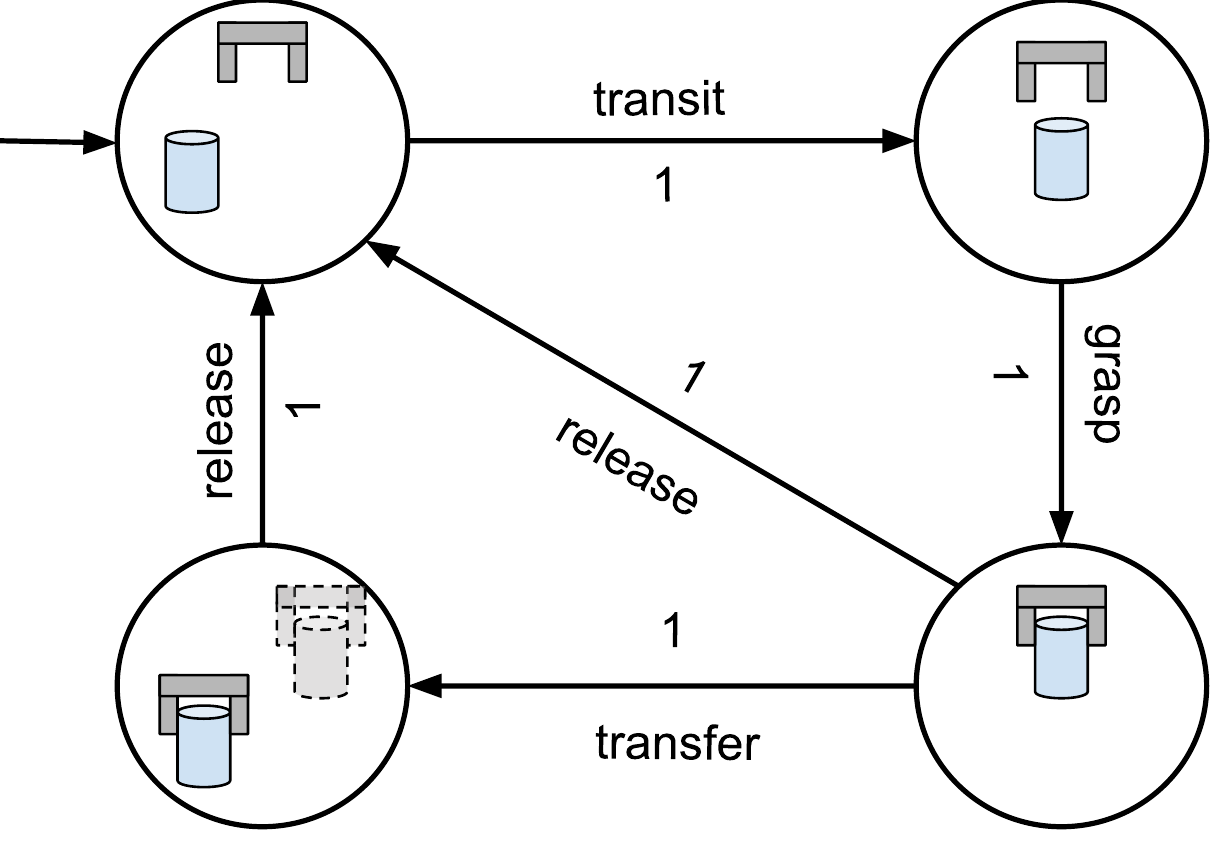}
        \caption{Transition System}
    \label{fig: mani_ts}
    \end{subfigure}%
    ~ 
    \begin{subfigure}[t]{0.45\columnwidth}
        \centering
          \includegraphics[width=0.99\linewidth]{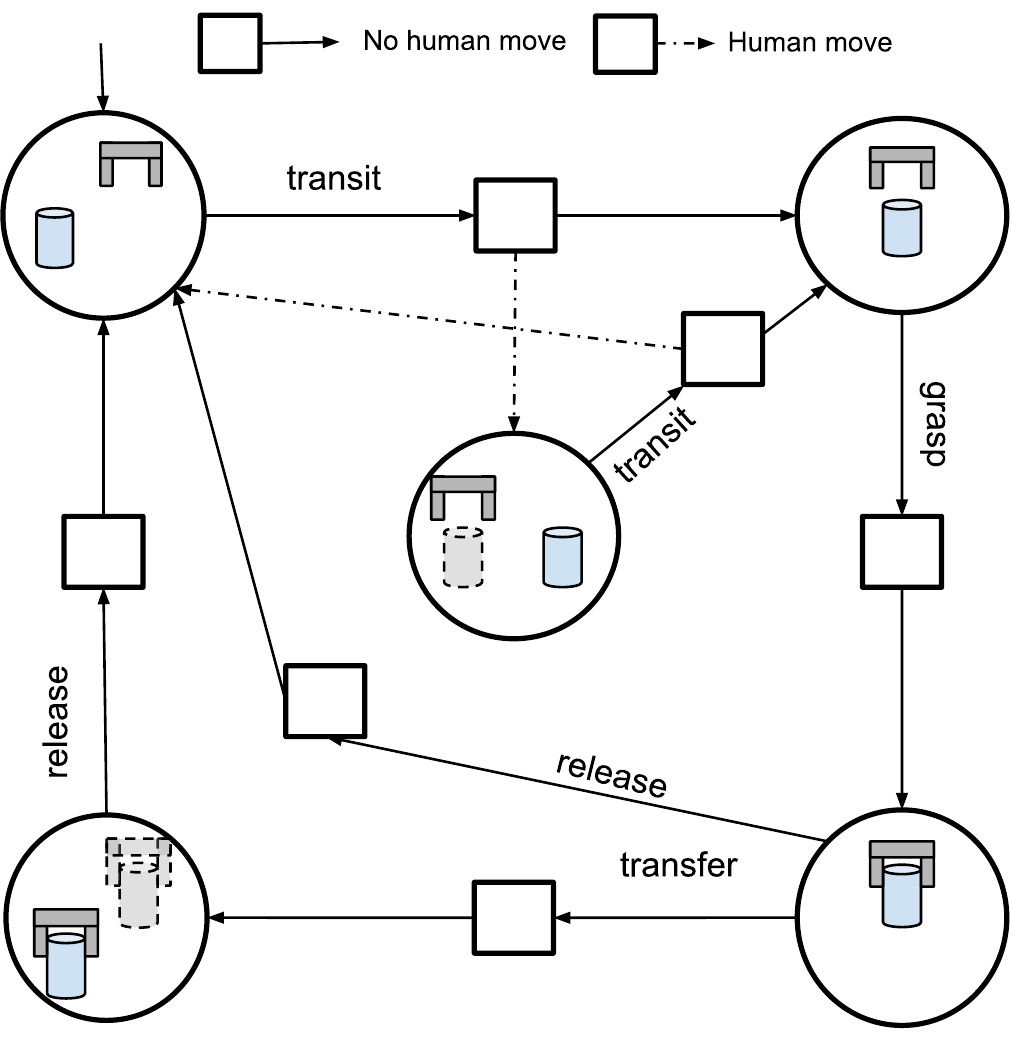}
        \caption{Two-player game}
    \label{fig: mani_game}
    \end{subfigure}%
\caption{\small Manipulation Domain $\G$ for Example \ref{ex: manip_domain}. States in circle and square belong to set $V_s$ and $V_e$, respectively.}
\label{fig: manip_domain}
% \vspace{-6mm}
\end{figure}

\else
\begin{figure}[t!]
    \centering
    \begin{subfigure}[t]{0.45\columnwidth}
        \centering
        % \resizebox{0.75\columnwidth}{!}{%
        % \includesvg[width=0.99\linewidth]{images/manip_domain_v2.svg}
        \includesvg[width=0.99\linewidth]{images/manip_domain_v4.svg}
        % }
        \caption{Transition System}
    \label{fig: mani_ts}
    \end{subfigure}%
    ~ 
    % \begin{subfigure}[t]{0.333\textwidth}
    \begin{subfigure}[t]{0.45\columnwidth}
        \centering
         % \resizebox{0.9\columnwidth}{!}{%
         % \includesvg[width=0.99\linewidth]{images/manip_domain_w_human_v1.svg}
          \includesvg[width=0.99\linewidth]{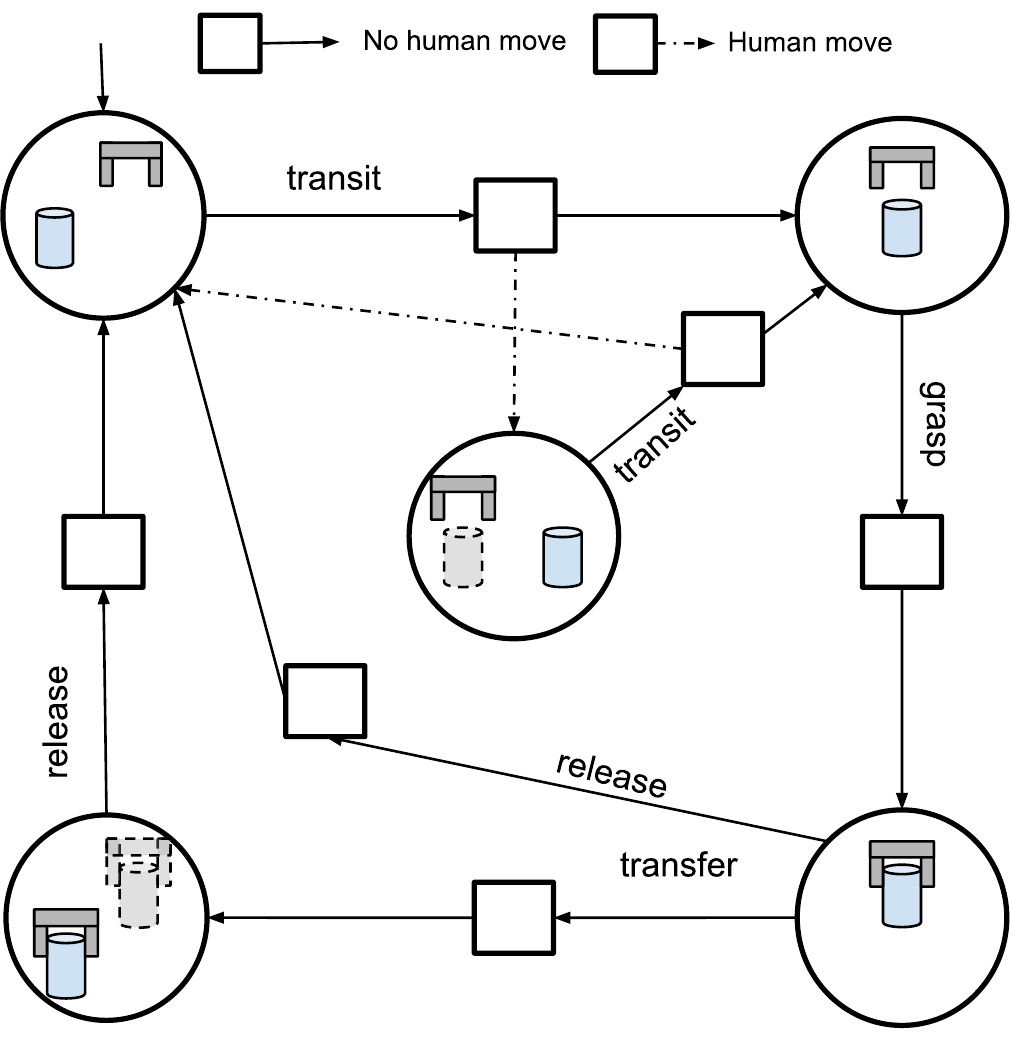}
         % }
        \caption{Two-player game}
    \label{fig: mani_game}
    \end{subfigure}%
\caption{\small Manipulation Domain $\G$ for Example \ref{ex: manip_domain}. States in circle and square belong to set $V_s$ and $V_e$, respectively.}
\label{fig: manip_domain}
% \vspace{-6mm}
\end{figure}
\fi

% To evolve over this game, starting at $v_0$, each player picks a strategy, which chooses the next action (an outgoing edge) given the sequence of states taken so far.  
% \ml{no need for the following sentence}
% At a robot state, the Robot chooses an action, and at the Env player's state, the Human chooses an action.

% \noindent
In this game, each player has a strategy to choose actions. Robot strategy $\sigma: V^* \cdot V_s \to A_s$ and human strategy $\tau: V^* \cdot V_e \to A_e$ pick an action for the respectively player given a finite sequence of states (history).  A strategy is called \textit{memoryless} if it only depends on the current state. The set of all strategies for the robot and human players are denoted by $\Sigma^\G$ and $\Tau^\G$, respectively.

% \begin{definition}[Strategy]
%     % \ml{need to be both formal and descriptive... e.g., A robot strategy $\sigma...$ chooses an action at a robot state given the sequence of states up to that point.  Similarly, a human strategy...}
%     A \textit{strategy} for the robot player denoted by $\sigma: V^* \cdot V_s \to A_s$ and human player denoted by $\tau: V^* \cdot V_e \to A_e$ picks an action for the respectively player given a finite sequence of states (history). 
%     % \ml{do you need to define the set of strategies for each player?}
%     A strategy is called \textit{memoryless} if it is only concerned with the current state, i.e., $\sigma(v_0\ldots v_n) = \sigma(v_n)$ if $v_n \in V_s$. 
%     The set of all strategies for the robot and human players are denoted by $\Sigma^\G$ and $\Tau^\G$, respectively.
% \end{definition}

% The unfolding of game $\G$ according to the players' strategies is a sequence of states called a play.

Given strategies $\sigma$ and $\tau$, the game unfolds, and a \textit{play} $\play (\sigma, \tau) =  v_0 v_1 \ldots v_n \in V^*$, which 
% \ml{$p$ is used for atomic proposition. created macro for play \textbackslash play}
% \km{I am using $z$ to denoted DFA states. Also, states in $\G$ are defined as $v$ so $\play(\cdot) = v_0 v_1 \ldots, \in V^*$, right?}
is a finite sequence of visited states, is obtained.
Intuitively, a play captures the evolution of the game (movement of the objects in the workspace) for some robot and human actions. 

Each action has a cost associated with it, e.g., energy.  We define the \textit{total payoff} $\Val^{v_0}(\play(\sigma,\tau))$ associated with a play to be the sum of the costs of the actions taken by the players, i.e., 
\begin{equation}
    \textstyle
    \Val^{v_0}(\play(\sigma, \tau)) := \sum_{i=0}^{n-1} F(v_i, a_i, v_{i+1}),
\end{equation}
where $a_i = \sigma(v_0 \ldots v_{i})$ if  $v_{i} \in V_s$, otherwise $a_i = \tau(v_0 \ldots v_{i})$.  
Further, a \textit{trace} for play $\play(\sigma, \tau)$ is the sequence of labels $\rho = L(v_0) L(v_1) \ldots L(v_n)$ of the states of $\play(\sigma, \tau)$.
Intuitively, the trace represents the evolution of the manipulation domain relative to a given task.

\subsection{LTL over finite Trace (LTLf)}

As most of robotic manipulation tasks are desired to be accomplished in finite time, a good choice to specify them unambiguously is LTLf \cite{vardi2013ltlf}. LTLf has the same syntax as Linear Temporal Logic (LTL), but the semantics are defined over finite executions (traces). 
% \ml{to save space, can reduce this to one sentence.}

\begin{definition}[LTLf Syntax \cite{vardi2013ltlf}] Given a set of atomic propositions $\Pi$, LTLf formula $\phi$ is defined recursively as:
    \begin{equation*}
        \phi := p \; | \neg \phi \;| \phi\; \wedge\; \phi \;| X\; \phi\; \;| \phi\, U \phi 
    \end{equation*}
    where $p \in \Pi$ is an atomic proposition, ``$\neg$" (negation) and $``\wedge$" (and) are boolean operators, and ``$X$" (next) and ``$U$" (until) are the temporal operators. 
    \label{def: ltlf_syntax}
\end{definition}
% \ml{in consistency in using Boolean vs boolean.... choose one and stick with it.}

\noindent
We define commonly used temporal operators ``$F$" (eventually) and ``$G$" (globally) as 
$F \phi := \top \; U\; \phi$ and $G \phi := \neg F \neg \phi$. 

The semantics of LTLf formulas is defined over finite traces in $\big(2^{\Pi} \big)^*$ \cite{vardi2013ltlf}.  
Notation $\rho \models \phi$ denotes that trace $\rho \in \big(2^{\Pi} \big)^*$ satisfies formula $\phi$.
We say that play $\play(\sigma,\tau)$
% induced by strategies $\sigma$ and $\tau$
accomplishes
% a task formula
$\phi$, denoted by $\play(\sigma,\tau) \models \phi$, if its trace $\rho \models \phi$.

\begin{example}
For our manipulation domain, 
% as shown in Fig. \ref{fig: manip_domain}, 
a task to build an arch with two support boxes and one colored box on top can be expressed as, 
$\phi_\text{arch} = F \left( p_{\text{box, support}_{1}} \land p_{\text{box, support}_{2}} \land p_\text{green, top} \right) \land 
    G \left( \neg(p_{\text{box, support}_{1}} \wedge p_{\text{box, support}_{2}}) \rightarrow \neg p_\text{green, top} \right).$
\end{example}
% ICRA exmaple
% For the arch-building example in Fig. \ref{fig: reg_illustration}, the task can be written as the LTLf formula
% \begin{align*}
%     \phi_\text{arch} =& F \left( p_\text{green, top} \wedge p_{\text{block, support}_{1}} \wedge p_{\text{block, support}_{2}} \right) \wedge \\ 
%     & G \left( \neg(p_{\text{block, support}_{1}} \wedge p_{\text{block, support}_{2}}) \rightarrow \neg p_\text{green, top} \right). \\
% \end{align*}
Our goal is to synthesize a strategy for the robot that can interact with the human to achieve its task $\phi$.  Specifically, we seek a strategy that guarantees the completion of $\phi$ by using no more than a given cost (energy) budget $\B \in \R_{\geq 0}$ under all possible human actions.  We refer to such a strategy as a winning strategy.  Formally, robot strategy $\sigma$ is called \textit{winning} if, for all $\tau \in \Tau^\G$, $\play(\sigma, \tau) \models \phi$ and $\Val(\sigma, \tau) \leq \B$.

% Given the manipulation domain as an abstraction $G$, the task $\phi$ specified as an LTLf formula, the objective reduces to synthesizing a strategy $\sigma$ for the robot that guarantees task completion. We define a strategy for the robot as a winning strategy $\sigma_{win}$ if following the strategy guarantees the robot can complete (satisfy) the given task for all possible human moves. We refer to the payoff associated with winning strategy $\sigma$, the minimum budget $\mathcal{B}$ needed by the robot to ensure satisfaction even in worst-case scenarios.

% \begin{definition}[Winning Strategy]
    % Given an LTLf task $\phi$ and a cost budget $\B \in \R_{\geq 0}$, robot strategy $\sigma$ is call winning if, for all $\tau \in \Tau^\G$, $p(\sigma, \tau) \models \phi$ and $\Val(\sigma, \tau) \leq \B$.    
% \end{definition}

% \subsection{Winning strategy}

The first problem we consider in this work is an efficient method of synthesizing a winning strategy.

\begin{problem}[Min-Max Reactive Synthesis]
    \label{problem1}
    Given 2-player manipulation domain game $\G$ and LTLf task formula $\phi$,
    % , and a Budget $\mathcal{B} \in \R_{\geq 0}$, 
    efficiently generate a winning strategy $\sigma^*$ 
    % exists and synthesize $\sigma_{win}$ if it is realizable.
    % and compute the budget required by the robot to enforce task completion.
    for the robot that guarantees the completion of task $\phi$ and minimizes its required budget $\B$ under all possible human strategies, i.e.,
    \begin{align*}
       \sigma^* = &\argmin_{\sigma \in \Sigma^\G} \max_{\tau \in \Tau^\G} \Val^{v_0}(\sigma, \tau) 
        \;\; \text{s.t. } \;  \play(\sigma, \tau) \models \phi \; \forall \tau \in \Tau^\G.
    \end{align*}
    % \qh{Uncomment below if too ugly}
    % \begin{align*}
    %    \sigma^* = &\argmin_{\sigma \in \Sigma^\G} \max_{\tau \in \Tau^\G} \Val^{v_0}(\play(\sigma, \tau)) \\
    %    & \text{s.t. }  \quad \play(\sigma, \tau) \models \phi  \qquad \qquad \forall \tau \in \Tau^\G.
    % \end{align*}
\end{problem}

Note that in the above formulation, the human is viewed as an adversary, i.e., it assumes human takes actions to maximize the cost budget of the robot. In the real world, however, humans are usually cooperative and willing to help.  Hence, the robot strategies obtained by the above formulation are typically ``unfriendly'' to the human, i.e., they seek to avoid interacting with the human.  
% This makes the interaction unproductive and unmeaningful for the human.  
Rather, we desire strategies that seek cooperation with the human while still considering that the human may not be fully cooperative.

% \subsection{Regret}

% Different notions of regret have been explored for games played on graphs \cite{raskin2016nonzero, guillermo2017reactive, filiot2010iterated, muvvala2021thesis}. In our previous approach \cite{muvvala2022regret}, we use regret to relax the adversarial assumption. This allows us to treat the human as a strategic agent with its own objective (possibly adversarial). 
 % Prior work has focused on finding the appropriate notion of regret and validating the emergent behavior on robotic manipulation tasks with possible collaboration with the human.
A formulation that enables such interaction is regret-based synthesis as shown in \cite{muvvala2022regret}.
% Intuitively, regret associated with a robot's action is a measure of how good that action is relative to the best action if the robot knew human's reaction in advance. 
Intuitively, regret is a measure of how good an action is relative to the best action if the robot knew the human's reaction in advance.

\begin{definition}[Regret \cite{filiot2010iterated}]
    Under robot and human strategies $\sigma$ and $\tau$, respectively, regret at state $v \in V$ is defined as 
    \begin{align}
            \reg^{v}(\sigma, \tau) = \Val^{v}(\sigma, \tau) - \min_{\sigma'}\, \Val^{v}(\sigma', \tau),
            \label{eq: reg_def}
    \end{align}
% where $\Val^{v}(\sigma, \tau)$ is the payoff under strategy $\sigma$ and $\tau$ from state $v$, and 
where $\sigma'$ is an alternate strategy to $\sigma$ $(\Sigma^{\G} \backslash \sigma )$ for the robot.
\end{definition}

% Formally, regret at state $v$ is defined as 
% \begin{align}
%         % \vspace{-2mm}
%         \reg^{v}(\sigma, \tau) = \Val^{v}(\sigma, \tau) - \min_{\sigma'}\, \Val^{v}(\sigma', \tau),
%         \label{eq: reg_def}
%         % \vspace{-2mm}
% \end{align}
% where $\Val^{v}(\sigma, \tau)$ is the payoff under strategy $\sigma$ and $\tau$ from state $v$, and $\sigma'$ is any alternate strategy for the robot other than $\sigma$. We use best-responses ($\min_{\sigma'}\, \Val^{v}(\sigma', \tau)$) to measure how good an action is for a fixed human strategy $\tau$. Hence, the  objective is to synthesize strategies for the robot that minimize regret, i.e., $\sigma^{\ast} = \arg \min_{\sigma} ( \max_\tau \reg^{v_0}(\sigma, \tau))$.

\noindent
This regret definition uses best-responses, i.e., $\min_{\sigma'}\, \Val^{v}(\sigma', \tau)$, to measure how good an action is for a fixed human strategy $\tau$. Hence, the objective is to synthesize strategies for the robot that minimize its regret under the assumption that human takes action to maximize robot's regret.

\begin{problem}[Regret-Minimizing Reactive Synthesis]
    \label{problem2}
    Given 2-player manipulation domain game $\G$, LTLf task formula $\phi$,
    and Budget $\B \in \R_{\geq 0}$, 
    efficiently compute a winning strategy $\sigma^*$ for the robot that guarantees not only the completion of task $\phi$ but also \emph{explores possible collaborations} with the human by minimizing its regret, i.e., 
    %  $$\sigma^* = \argmin_{\sigma \in \Sigma^\G} \max_{\tau \in \Tau^\G} \reg^{v_0}(\sigma, \tau)$$
    % such that  $\play(\sigma, \tau) \models \phi, \;\,  \;\, \Val^{v_0}(\sigma,\tau) \leq \B  \qquad \forall \tau \in \Tau^{\G}.$
    % \begin{align*}
    %    \sigma^* = &\argmin_{\sigma \in \Sigma^\G} \max_{\tau \in \Tau^\G} \reg^{v_0}(\sigma, \tau) \\
    %    & \text{s.t. }  \quad \play(\sigma, \tau) \models \phi, \;\,  \;\, \Val^{v_0}(\sigma,\tau) \leq \B  \qquad \forall \tau \in \Tau^{\G}.
    % \end{align*}
    \begin{align*}
       \sigma^* = &\argmin_{\sigma \in \Sigma^\G} \max_{\tau \in \Tau^\G} \reg^{v_0}(\sigma, \tau) \\
       & \text{s.t. }  \quad \play(\sigma, \tau) \models \phi  \qquad \qquad \quad \;\, \forall \tau \in \Tau^\G \\
       & \qquad \;\, \Val^{v_0}(\sigma,\tau) \leq \B  \qquad \qquad \forall \tau \in \Tau^{\G}.
    \end{align*}
\end{problem}

There are generally three major challenges in the above problems.  
% (i) Manipulation domains are notoriously known to suffer from the \textit{state-explosion} problem, i.e., given a set $L$ of locations and set $O$ of objects, the size of the abstraction is $\mathcal{O}(|L|^{|O|})$.  Here, this challenge is exacerbated by considering an extension of the domain to human-robot manipulation games as well as high-level complex tasks, requiring composition of the game domain with the task space.  (ii) These games are quantitative, requiring to reason not only about the completion of the task but also the needed quantity (budget), which adds numerical computations to the already-expensive qualitative algorithms.  (iii) While memoryless strategies are sufficient for Problem~\ref{problem1}, regret minimizing strategies (Problem~\ref{problem2}) are necessarily finite memory, making the computations even more challenging \cite{muvvala2022regret}. 
\begin{enumerate}
    \item Manipulation domains are notoriously known to suffer from the \textit{state-explosion} problem, i.e., given a set $L$ of locations and set $O$ of objects, the size of the abstraction is $\mathcal{O}(|L|^{|O|})$.  Here, this challenge is exacerbated by considering an extension of the domain to human-robot manipulation games as well as high-level complex tasks, requiring composition of the game domain with the task space.  
    \item These games are quantitative, requiring to reason not only about the completion of the task but also the needed quantity (budget), which adds numerical computations to the already-expensive qualitative algorithms.  
    \item While memoryless strategies are sufficient for Problem~\ref{problem1}, regret minimizing strategies (Problem~\ref{problem2}) are necessarily finite memory \cite{muvvala2022regret}, making the computations even more challenging. 
\end{enumerate}
% \ml{to save space, remove the enumerate env above and add numbers in a paragraph.}

Existing algorithms for Problems~\ref{problem1} and \ref{problem2} are based on explicit construction of the game venue's graph. 
%states and edges
They are hence severely limited in their scalability and cannot find solutions to real-world manipulation problems in a reasonable amount of time.  
In addition, for regret-minimizing strategies, the algorithms require large amount of memory (tens of GB) even for small problem instances (see benchmarks in \cite{muvvala2022regret}).  
To overcome these challenges, rather than relying on explicit construction, we base our approach on symbolic methods.  
% We specifically use Boolean variables and operations to construct symbolic games.  We then develop algorithms for these symbolic games to approach Problems~\ref{problem1} and \ref{problem2}.  

% The existing approach builds a product graph by composing the Manipulation abstraction $\mathcal{G}$ and DFA $\mathcal{A}_\phi$ to construct DFA Game $\PA$. Intuitively, this game captures all possible ways in which the Robot can achieve task $\phi$ for all possible human interventions. We then synthesize regret-minimizing strategies in line with Problem \ref{problem} from \cite{muvvala2022regret}. 

% explicitly build the graph by storing the states and edges in the memory. 
\section{Preliminaries}
\label{sec: prelimaries}

% In this section, we briefly introduce the preliminaries necessary for our framework. We seek to synthesize regret-minimizing strategies efficiently. Thus, we introduce symbolic graphs that are commonly used in practice for systems with large state space. Reasoning over such models explicitly becomes computationally expensive and time-consuming. We then proceed to discuss how we reformulate our algorithms to accommodate symbolic data structures and scale up our implementation
% to reason over more objects with complex tasks
% using symbolic techniques.

In this section, we briefly introduce the preliminaries needed for our framework.  First, we review the explicit algorithm for quantitative games that solves Problem~\ref{problem1}, and then present BDDs, which are structures used to represent boolean functions.  
% \ml{to save space, can shorten this paragraph}

\subsection{Overview of Explicit Strategy Synthesis}

\textbf{DFA: }
% First, we convert the task $\phi$ into a Deterministic Finite Automata (DFA) $\mathcal{A}_\phi$. Next, we compose the Game abstraction $G$ and the $\mathcal{A}_\phi$ to construct a DFA Game $\PA$. This game captures all possible ways in which the human and the robot can interact and can accomplish the task. 
The approach proposed in \cite{he2017reactive} for Problem \ref{problem1} is to first construct a Deterministic Finite Automata (DFA) for a given task $\phi$. A DFA is defined as a tuple $\mathcal{A}_\phi = (Z, z_0, 2^\Pi, \delta_{\phi}, Z_f)$, where $Z$ is a finite set of states,  $z_0$ is the initial state, $2^\Pi$ is the alphabet, $\delta_{\phi}: Z \times 2^\Pi \to Z$ is the deterministic transition function, and $Z_f \subseteq Z$ is the set of accepting states. 
% For every LTLf formula $\phi$, a DFA $\mathcal{A}_\phi$ can be constructed that accepts precisely the set of traces that satisfies $\phi$ \cite{vardi2013ltlf}.
% Given task $\phi$, we construct DFA $\mathcal{A}_\phi = (Z, z_0, \Sigma, \delta, Z_f)$, where $Z$ is a finite set of states,  $z_0$ is the initial state, $\Sigma = 2^\Pi$ is the alphabet, 
% % , Boolean properties that are relevant to task $\phi$, 
% $\delta: Z \times \Sigma \rightarrow Z$ is the deterministic transition function, and $Z_f \subseteq Z$ is the set of accepting states.
A run of $\mathcal{A}_\phi$ on a trace $\rho = \rho[1] \rho[2] \ldots \rho[n]$, where $\rho[i] \in 2^{\Pi}$, is a sequence of DFA states $z_0 z_1 \ldots z_{n}$, where $z_{i+1} = \delta_{\phi}(z_i, \rho[i+1])$ for all $0\leq i \leq n-1$. 
If $z_n \in Z_f$, the path is called accepting, and its corresponding trace $\rho$ is accepted by the DFA.
The DFA $\mathcal{A}_\phi$ reasons over the task $\phi$ and exactly accepts traces that satisfy $\phi$ \cite{vardi2013ltlf}.  Thus, $\mathcal{A}_\phi$ captures all possible ways of accomplishing $\phi$. 

\textbf{DFA Game: }
Next, the algorithm composes the game abstraction $\G$ with $\mathcal{A}_\phi$ to construct a DFA game $\PA = \G \times \mathcal{A}_\phi$. This DFA game is a tuple $\PA~=~(S, S_f, s_0, A_s, A_e, F, \delta_\PA)$, where $A_s$, $A_e$, and $F$ are as in Def.~\ref{def:game abstraction}, $S = S_s \cup S_e$ is a set of states with $S_s = V_s\times Z$ and $S_e = V_e\times Z$, $S_f = V \times Z_f$ is the set of accepting states, $s_0=(v_0, z_0)$ is the initial state, and $\delta_\PA$ is a transition function such that $s' = \delta_{\PA}(s, a)$, where $s = (v, z)$, $s' = (v', z')$, if $v' = \delta(v,a)$ $a$ and $z' = \delta_\phi(z,L(v'))$.
% that models all possible ways in which the robot can interact with the human to accomplish task $\phi$. 
Intuitively, $\PA$ augments each state in $\G$ with the DFA state $z$ as a ``memory mode" to keep track of the progress made towards achieving task $\phi$. The DFA game evolves by first evolving over the manipulation domain $\G$ and then checking the atomic propositions that are true at the current state and update the DFA state accordingly. 
% Each state $s= (v, z)$ belongs either to the Sys player, if $v \in V_s$, or else to the Env player. Thus, the set of states $S = S_s \cup S_e$ and $S_s \cap S_e = \emptyset$. The set of states $s \in S_f$ are called accepting states if the corresponding DFA state is an accepting state, i.e., $z \in Z_f$. 
% The edge weights associated with edge  in $\PA$ are determined by $\delta_{\PA}(s, a, s')$ where $s = (v, z)$, $s' = (v', z')$ and $a$ is a valid action as per $\delta(v, a, v')$.

\textbf{Strategy Synthesis: }
The classical approach for synthesizing winning strategies $\sigma^*$ on $\PA$ is to play a min-max reachability game \cite{gimbert2004can, brihaye2017pseudopolynomial}. Given a set of accepting states $S_f$, we define operators, $\cPreMax(s)$ and $\cPreMin(s)$, that update the state values as follows.
\begin{align}
    \label{eq: pre_op_max}
    \cPreMax(s) & = \max_{a}(F(s, a) + W(s')) \quad \text{if } s \in S_e, \\
    \cPreMin(s) & = \min_{a}(F(s, a) + W(s’)) \quad \text{if } s \in S_s,
    \label{eq: pre_op_min}
\end{align}
% \ml{shouldn't min/max be over $a \in A_e / A_s$ and $s' = \delta(s,a)$? If so, also need to fix the alg.}
% \ml{missing actions?  Also, if it's pre, then shouldn't it be $\cPreMax(s) = \max_{s'}(F(s', s) + W(s))...$? }
where 
$s' = \delta(s, a)$, 
$W$ is a vector of state values, and $W(s')$ is the value associated with state $s'$. 
Then, the algorithm 
% \qh{can remove "for ... constraints"}
% for games with reachability objective under quantitative constraints
for reachability games with quantitative constraints
is to compute the (least) fixed point of the operators \eqref{eq: pre_op_max} and \eqref{eq: pre_op_min} by applying the operator over all the states in $\PA$. Thus, playing a min-max reachability game reduces to fixed-point computation and is summarized in Alg.~\ref{algo: exp_value_iteration} \cite{ brihaye2017pseudopolynomial}. While the value-iteration algorithm for min-max reachability is polynomial
% \ml{remind me again, why is it pseudo-polynomial?  I thought regret games were pseudo-poly.}
\cite{gimbert2004can}, fixed-point computation over large graphs ($|S|$ is very large) becomes computationally expensive in terms of time and memory. 
To mitigate this issue, symbolic graphs are often employed where the state and transition function are represented as boolean functions. 
Below, we discuss how to store and manipulate the boolean functions.
% \ml{don't we need to initialize all the weights with $\infty$ except the ones for the accepting states?}

% {\fontsize{9}{9}\selectfont
\setlength{\floatsep}{2pt}% Remove \textfloatsep
\setlength{\textfloatsep}{2pt}% Remove \textfloatsep
\begin{algorithm}[tb]
    \caption{Explicit Value Iteration}
    \label{algo: exp_value_iteration}
    \SetKwInOut{Input}{Input}\SetKwInOut{Output}{Output}
    
    \SetKwData{sval}{sval}
    \SetKwData{Win}{Win}
    \SetKwData{aPre}{aPre}
    \SetKwData{uPre}{uPre}

    \SetKwData{Strategy}{Strategy}
    \SetKwData{hStrategy}{hStrategy}

    %  ADD and BDD operations
    \SetKwFunction{addZero}{addZero}
    \SetKwFunction{addInfinity}{addInfinity}
    \SetKwFunction{restrict}{restrict}
    \SetKwFunction{min}{min}
    \SetKwFunction{max}{max}
    \SetKwFunction{argmax}{argmax}
    \SetKwFunction{argmin}{argmin} 
    \SetKwFunction{add}{add}
    \SetKwFunction{ite}{ite}
    \SetKwFunction{preimage}{preimage}
    \SetKwData{This}{this}\SetKwData{Up}{up}
    \SetKwFunction{ValueIteration}{ValueIteration}   
    \Input{Explicit DFA Game $\PA$}
    \Output{Winning strategy $\sigma_{win}$, Optimal values $W$}
    % initial set of winning state to be zero
    % $W_F := 0$ \tcc*[r]{set of target states} 
    $W(S) \leftarrow \infty; \quad \sigma \leftarrow \emptyset; \quad \tau \leftarrow \emptyset; \quad W' \leftarrow \emptyset$ \\
    $W(S_f) \leftarrow$ \text{initialize all accepting state values as zero}\\
    \While{$W' \neq W$}{
        $W' = W$ \\
        $W(s) = \max(F(s, a) + W'(s’)) \quad \forall s \in S_e$ \\
        $\tau_s = \argmax_{a}(F(s, a) + W'(s’)) \quad \forall s \in S_e$ \\
        $W(s) = \min(F(s, a) + W'(s’)) \quad \forall s \in S_s$ \\
        $\sigma_s = \argmin_{a}(F(s, a) + W'(s’)) \quad \forall s \in S_s$ \\
    } 
    \KwRet{$W$, $\sigma$}
\end{algorithm}
% }

% Binary Decision Diagram is a common used symbolic data structure to efficient reprersent and manipulate these boolean formulas.

% \km{talk about value iteration. At the heart is fixed point computation, which depend on set based operation and predecessor computation. These operations are primitive operations that are ready available. Talk about complexity of these operation here. Equiavalence checking - constant time, quantification - quadratic, etc..}

\subsection{Symbolic Data Structures:  BDDs and ADDs}

Symbolic Data structures like BDDs \cite{somenzi1999binary, bryant1995binary} have been extensively used by the model-checking and verification community due to their ability to perform fixed-point computations over large state spaces efficiently \cite{burch1992symbolic, chaki2018bdd}.
% \ml{to save space, the above sentence can be removed.}
BDD is a canonical directed-acyclic graph representation of a boolean function $f: 2^{X} \to \{0, 1\}$ where $X$ is a set of boolean variables. To evaluate $f$, we start from the root node and recursively evaluate the boolean variable $x_i \in X$ at the current node until we reach the terminal node. 
The order in which the variables appear along each path is fixed and is called variable ordering. Due to the fixed variable ordering, BDDs are canonical, i.e., the same BDD always represents the same function.
The power of BDDs is that they allow compact representation of functions like transition relations and efficiently perform set 
operations, e.g., conjunction, disjunction, negation. 

\begin{figure*}[t!]
    \centering
    \scalebox{.73}{
    \begin{subfigure}[t]{0.3\textwidth}
        \centering
        \resizebox{\columnwidth}{!}{%
         \begin{tikzpicture}[shorten >=1pt,node distance=6cm,>=stealth',thick, auto,
                        text style/.style={sloped, text=black, font=\large, above},
                        every state/.style={fill,very thick,black!20,text=black,\shadowString},
                        human/.style = {fill,very thick,black!20,rounded corners, text=black, shape=rectangle, minimum height=1cm,minimum width=1cm, blur shadow},
                        accepting/.style ={blue!50!black!50,text=white,accepting by double},
                        initial/.style ={red!80!black!40,text=black,initial by arrow, initial above}, initial text=$ $]
                        
        \node[state, initial, label=above right:\texttt{$\{ready, p_{00}\}$}] (r1) {$v_{s1}$};
        \node[state, label=above left:\texttt{$\{to-obj, p_{00}\}$}] at (8, 0) (r2) {$v_{s2}$};
        \node[state, label=below right:\texttt{$\{ready', p_{00}\}$}] at (4, -4) (r3) {$v_{s3}$};
        \node[state, label=below left:\texttt{$\{hold, p_{ee}\}$}] at (8, -8) (r4) {$v_{s4}$};
        \node[state, label=above right:\texttt{$\{to-loc, p_{ee}\}$}] at (0, -8) (r5) {$v_{s5}$};
        
        \path[->]
            (r1) edge node[text style, above] {\texttt{ts \& no human-move }} (r2)
            (r1) edge node[text style, below] {1} (r2)
            (r2) edge node[text style, above] {\texttt{gr \& no human-move}} (r4)
            (r2) edge node[text style, below] {1} (r4)
            (r4) edge node[text style, above] {\texttt{tr \& no human-move}} (r5)
            (r4) edge node[text style, below] {1} (r5)
            (r4) edge[bend left, pos=0.5] node[text style, below] {\texttt{rl \& no human-move}} (r1)
            (r4) edge[bend left, pos=0.5] node[text style, above] {1} (r1)
            (r5) edge node[text style, above] {\texttt{rl \& no human-move}} (r1)
            (r5) edge node[text style, below] {1} (r1)
            (r3) edge[bend left, pos=0.5, dash pattern = on 1 pt off 2 pt on 3 pt off 2 pt] node[text style, below] {} (r1)
            (r3) edge[bend left, pos=0.5, dash pattern = on 1 pt off 2 pt on 3 pt off 2 pt] node[text style, above] {1} (r1)
            (r1) edge[bend left, pos=0.5, dash pattern = on 1 pt off 2 pt on 3 pt off 2 pt] node[text style, below] {1} (r3)
            (r3) edge node[text style, above] {{\texttt{ts \& no human-move }}} (r2)
            (r3) edge node[text style, below] {1} (r2);
    \end{tikzpicture}
    }
    \caption{\small Explicit Two-player game abstraction}
    \label{fig: exp_no_human_mani_game}
    \end{subfigure}}%
    ~
    \scalebox{.78}{
    \begin{subfigure}[t]{0.4\textwidth}
        \centering
        \resizebox{0.75\columnwidth}{!}{%
         \begin{tikzpicture}[shorten >=1pt,node distance=6cm,>=stealth',thick, auto,
                        % text style/.style={sloped, text=black, font=\footnotesize, above},
                        text style/.style={sloped, text=black, font=\large, above},
                        every state/.style={fill,very thick,black!20,text=black,\shadowString},
                        human/.style = {fill,very thick,black!20,rounded corners, text=black, shape=rectangle, minimum height=1cm,minimum width=1cm, blur shadow},
                        accepting/.style ={blue!50!black!50,text=white,accepting by double},
                        initial/.style ={red!80!black!40,text=black,initial by arrow, initial above}, initial text=$ $]
                        
        \node[state, initial] (r1) {\;\; $x_0 \wedge x_1 \wedge x_2$ \;\;};
        \node[state] at (6, 0) (r2) {\; $x_0 \wedge \neg x_1 \wedge x_2$ \;};
        \node[state] at (6, -6) (r4) {$\neg  x_0 \wedge x_1 \wedge \neg x_2$ \;};
        \node[state] at (0, -6) (r5) {$\neg  x_0 \wedge \neg x_1 \wedge \neg x_2$};
        \path[->]
            (r1) edge node[text style, above] {$o_0 \wedge o_1 \wedge \neg i_0$} (r2)
            (r2) edge node[text style, above] {$o_0 \wedge \neg o_1 \wedge \neg i_0$} (r4)
            (r4) edge node[text style, above] {$\neg o_0 \wedge o_1 \wedge \neg i_0$} (r5)
            (r4) edge node[text style, below] {$\neg o_0 \wedge \neg o_1 \wedge \neg i_0$} (r1)
            (r5) edge node[text style, above] {$\neg o_0 \wedge \neg o_1 \wedge \neg i_0$} (r1);
    \end{tikzpicture}
    }
    \caption{\small Symbolic Two-player game}
    \label{fig: boolean_mani_game}
    \end{subfigure}}%
    ~~~
    \scalebox{.78}{
    \begin{subfigure}[t]{0.15\textwidth}
        % \begin{adjustbox}{width=\linewidth}
        \centering
        \resizebox{\columnwidth}{!}{%
        \begin{tikzpicture}[shorten >=1pt,node distance=6cm,>=stealth',thick, auto,
                            every state/.style={fill,very thick,black!20,text=black},
                            terminal/.style = {fill,very thick,black!20,rounded corners, text=black, shape=rectangle, minimum height=1cm,minimum width=1cm},
                            accepting/.style ={blue!50!black!50,text=white,accepting by double},
                            initial/.style ={red!80!black!40,text=black,initial by arrow, initial above}, initial text=$ $]
            
            \node[state, initial] (x0) {$x_0$};
            \node[state] (x1) at (-1, -1.75) {$x_1$};
            \node[state] (s_x1) at (1, -1.75) {$x_1$};
            \node[state] (x2) at (-0.35, -3) {$x_2$};
            \node[state] (s_x2) at (1, -3) {$x_2$};
            \node[terminal] (true) at (1, -5) {1};
            \node[terminal] (false) at (-1, -5) {0};
            
            \path[-]
                (x0) edge [dotted] (x1)
                (x0) edge (s_x1)
                (x1) edge [dotted] (false)
                (s_x1) edge [dotted] (false)
                (x1) edge (x2)
                (s_x1) edge (s_x2)
                (x2) edge [dotted] (true)
                (x2) edge (false)
                (s_x2) edge [dotted] (false)
                (s_x2) edge (true);
        \end{tikzpicture}
        }
        % \end{adjustbox}
    \caption{\small BDD for set $V'$}
    \label{fig: bdd_ex}
    \end{subfigure}}%
    ~~~~~~~
    \scalebox{.85}{
    \begin{subfigure}[t]{0.15\textwidth}
        \centering
        % \begin{adjustbox}{width=\linewidth}
        \resizebox{\columnwidth}{!}{%
        \begin{tikzpicture}[shorten >=1pt,node distance=6cm,>=stealth',thick, auto,
                            every state/.style={fill,very thick,black!20,text=black},
                            terminal/.style = {fill,very thick,black!20,rounded corners, text=black, shape=rectangle, minimum height=1cm,minimum width=1cm},
                            accepting/.style ={blue!50!black!50,text=white,accepting by double},
                            initial/.style ={red!80!black!40,text=black,initial by arrow, initial above}, initial text=$ $]
            
            \node[state, initial] (x0) {$x_0$};
            \node[state] (x1) at (1.2, -1.75) {$x_1$};
            \node[state] (s_x1) at (-1.2, -1.75) {$x_1$};
            \node[state] (x2) at (1.2, -3) {$x_2$};
            \node[state] (s_x2) at (-1.2, -3) {$x_2$};
            \node[terminal] (0) at (0, -5) {0};
            \node[terminal] (2) at (1.2, -5) {2};
            \node[terminal] (3) at (-1.2, -5) {3};
            
            \path[-]
                (x0) edge [dotted] (s_x1)
                (x0) edge (x1)
                (x1) edge (0)
                (x1) edge [dotted] (x2)
                (s_x1) edge [dotted] (s_x2)
                (s_x1) edge (0)
                (s_x2) edge [dotted] (3)
                (s_x2) edge (0)
                (x2) edge [dotted] (0)
                (x2) edge (2)
                ;
        \end{tikzpicture}
        % \end{adjustbox}
        }
    \caption{\small ADD for set $V''$ %\km{Update it!}
    }
    \label{fig: add_ex}
    \end{subfigure}}%
\caption{\small (a) and (b) illustrate explicit graph and its equivalent symbolic graph, respectively. Robot actions \texttt{transit} (ts) is $o_0 \wedge o_1$,  \texttt{grasp} (gr) is $o_0 \wedge \neg o_1$,  \texttt{transfer} (tr) is $\neg o_0 \wedge o_1$, and  \texttt{release} (rl) is $\neg o_0 \wedge \neg o_1 $. Human actions  \texttt{human-move} (dashed) and  \texttt{no human-move} (solid) are $i_0$ and $\neg i_0$, respectively. We skip \texttt{human-move} edges for simplicity. Variables $x_0, x_1 \in X$ correspond to robot configuration  and $x_2 \in X$ for box being grounded ($p_{00}$) or manipulated ($p_{ee}$). (c) and (d)
illustrate the BDD and ADD for Example \ref{ex: bdd_ex} and \ref{ex: add_ex}.
% \ml{fix this caption...}
}
% We represent Robot configuration propositions with $x_0, x_1 \in X$ and objects proposition grounded ($p_{00}$) or manipulated ($p_{ee}$) with $x_2 \in X$.}
% The realization of the boolean function $f$ over the BDD evaluates to true (1) or false (0) while in the ADD, the realization evaluates to 2 and 3 for $v_{s2}$ and $v_{s3}$, zero otherwise.
% Note we do not need to evaluate all boolean variables to check if the boolean formula associated with the path is true or false.}
\label{fig: manip_and_dd_ex}
% \vspace{-4mm}
\end{figure*}

% \caption{Explicit Graph to Symbolic Graph with Boolean representation. We skip evolution to state $v_{s4}$ from Fig. \ref{fig: mani_game_graph} for clarity. Robot actions \texttt{transit} is $o_0 \wedge o_1$,  \texttt{grasp} is $o_0 \wedge \neg o_1$,  \texttt{transfer} is $\neg o_0 \wedge o_1$, and  \texttt{release} is $\neg o_0 \wedge \neg o_1 $. Human action  \texttt{human-move} and  \texttt{no human-move} are $i_0$ and $\neg i_0$, respectively. We represent Robot configuration propositions like ready, hold, to-loc, and to-obj with $x_0, x_1 \in X$ and proposition relevant to the box being grounded ($p_{00}$) or manipulated ($p_{ee}$) with $x_2 \in X$.}

% \label{fig: manip_domain}
% \end{figure*}

%available font sizes
% \tiny
% \scriptsize
% \footnotesize
% \small
% \normalsize
% \large
% \Large
% \LARGE
% \huge
% \Huge

\begin{example}
% \ml{Move Fig. \ref{fig: manip_domain} here and fit both within the column-width}
 Consider the explicit Graph $\G$ in Fig. \ref{fig: exp_no_human_mani_game}. Denote each state $v \in V$ by its corresponding boolean formula over boolean variables $X = \{x_0, x_1, x_2\}$. This is shown in Fig. \ref{fig: boolean_mani_game}. Fig. \ref{fig: bdd_ex} shows the BDD corresponding to boolean function representing subset of states $V' = \{v_{s1}, v_{s4}\}$, i.e, $f(x_0, x_1, x_2) = (x_0 \wedge x_1 \wedge x_2) \vee (\neg x_0 \wedge x_1 \wedge \neg x_2)$. 
 % \km{Change this to be over state labels}
\label{ex: bdd_ex}
\end{example}

% BDDs have a number of useful properties, including compactness (they can represent many functions like Transition Relations using a relatively small number of nodes), Canonicity, and efficient operations (e.g., conjunction, disjunction, negation). 

Algebraic Decision Diagrams (ADDs), also known as Multi-Terminal BDDs, are a generalization of BDDs, where the realization of a boolean formula can evaluate to some set \cite{bahar1997algebric}. Mathematically, ADDs represent boolean function $f: 2^{X} \to C$, where $X$ is a set of boolean variables and $C \subset \R$ is a finite set of constants. 
% For our framework $C \subset \mathbb{R}$. 
Similar to BDDs, ADDs are also built on principles of Boole's Expansion  Theorem \cite{boole1847mathematical}. Thus, all the binary operations that are applicable to BDDs are translated to ADDs as well.
\begin{example}
% Fig. \ref{fig: add_ex}  shows Boolean function for states $v_{s2}$, and $v_{s3}$ whose realization is 2 and 3, respectively
 % Similarly, say
 We assign, to states in $V'' = \{v_{s2}, v_{s5}\}$ in Fig. \ref{fig: exp_no_human_mani_game}, integer values 2 and 3, respectively. The corresponding ADD for $f(x_0, x_1, x_2) = \left((x_0 \wedge \neg x_1 \wedge x_2) \to 2 \right)  \vee \left((\neg x_0 \wedge \neg x_1 \wedge \neg x_2) \to 3 \right)$ is shown in Fig. \ref{fig: add_ex}.
\label{ex: add_ex}
 % \km{Change this to be over robot actions}
\end{example}

% Canonicity enables BDDs to perform binary and set-based operations such as in constant or linear time \km{add citation here}. This makes BDD an ideal choice for our problem as at the core of the Reachability-based approaches is the computation of Pre-images, set equivalence checking and quantification.

\section{Symbolic Games}
\label{sec: approach}

% It is well known that conversion from LTLf formula $\phi$ to DFA $\mathcal{A}_\phi$ is doubly exponential \cite{vardi2013ltlf}, and Regret-minimizing strategy synthesis is Pseudo-polynomial in the size of $\PA$ \cite{filiot2010iterated}. 
% This can render the strategy synthesis process intractable or even impossible due to limitations in computing power and time.
% Thus, the existing approach suffers from the state explosion problem.
% To mitigate this issue, we propose using a symbolic approach that scales better than the explicit approach and has faster computation times. 
% To mitigate these problems, we base our approach on symbolic methods.

% To achieve a scalable and efficient synthesis framework, we base our approach on symbolic methods.
Here, we introduce symbolic graphs, their construction, and our proposed algorithms to perform symbolic Value Iteration (VI) and synthesize regret-minimizing strategies in a symbolic fashion.

\subsection{Symbolic Two-player Games}

To avoid explicit construction of all the states and transitions, we use a symbolic representation that lets us reason over sets of states and edges. Usually, using this approach results in a much more concise representation than explicitly representing each state and edge. However, we need to redefine algorithms to accommodate symbolic representations. 

% Specifically, each state is represented as a conjunction of boolean variables. For Monolithinc Encoding, we have a second pair of boolean variables that represent the next state. Thus, a valid edge on the Explicit graph can then be represented as the conjunction of formulas associated with current and next state variables. Hence, every formula corresponding to state (and edge) is evaluated to 1.

Encoding the game graph $\G$ using boolean formulas enables us to represent the states and actions using only a logarithmic number of boolean variables. Further, we can use BDDs (ADDs) to compactly represent these boolean formulas and perform efficient set-based operations using operations defined over BDDs (ADDs). 

\begin{definition}[Compositional Symbolic Game $\G_s$ \cite{zhu2017symbolic}]
Given a Two-player game $\G$, its corresponding symbolic representation is defined as tuple $\G_s = (X, I, O, \eta)$ where,
\begin{itemize}
    \item $X = \{x_0,x_1,\ldots, x_n\}$
    is the set of boolean variables such that every state $v \in V$ has corresponding boolean formula whose realization is true, 
    % and $|X| = Log (|V|)$,  %i.e., $2^{X} \to 1$, DONT NEED IT
    \item $I$ is the set of boolean variables such that every human action $a_e \in A_e$ has corresponding boolean formula whose realization is true, %i.e., $2^{I} \to 1$,
    \item $O$ is the set of boolean variables such that every robot action $a_s \in A_s$ has corresponding boolean formula whose realization is true, 
    % i.e., $2^{O} \to 1$, 
    and
    % \item $X$, $I$, $O$ have the same definition as Definition \ref{def: mono_sym_game}
    \item transition relation $\eta = \{\eta_{x_0}, \eta_{x_1}, \ldots, \eta_{x_n} \}$ 
    % \ml{still infinite set... check all}
    % is a vector of boolean functions where $\eta_{x_i}: 2^{X} \times 2^{I} \times 2^{O} \to 1$
    % \ml{is $x_i$ a set?}
    % , i.e.,  $\eta_{x_i}(X, I, O)$ evaluates to true 
    % \ml{hhmm.... so the range of $\eta$ is $\{0,1\}$, no?}
    % iff $x_i \in X$ is true in the corresponding successor state, \km{and $|\eta| = |X|$}.
    % \ml{how about this:
    is a set of boolean functions, where $\eta_{x_i}: 2^{X} \times 2^{I} \times 2^{O} \to \{0,1\}$
    such that $\eta_{x_i}(X, I, O) = 1$ if $x_i \in X$ is true in the corresponding successor state, otherwise 0.
    % }
    
\end{itemize}
\label{def: comp_sym_game}
\end{definition}

\begin{remark}
    Since our interest is in synthesizing a robot strategy, we do not need to compute $W$ over human states explicitly. Hence, we abstract away the states in $V_e$ by modifying the transitions in $\G$ to model the evolution from each robot state to be a function of robot action $a_s$ and human action $a_e$, i.e., $\delta: V_s \times A_s \times A_e \to V_s$ for all $v \in V_s$. Fig. \ref{fig: exp_no_human_mani_game} depicts the modified abstraction for the abstraction in Fig. \ref{fig: mani_game}.
    % \ref{fig: mani_game_graph}.
    % \ml{Where are the nondeterministic outcomes due to human actions in Fig. \ref{fig: exp_no_human_mani_game}?}
\end{remark}

% We can infer from Compositional encoding's definition that we need fewer boolean variables than Monolithic encoding. Further, previous works \cite{he2019symbolic, zhu2017symbolic} have shown that using Compositional encoding is computationally less expensive and more efficient in terms of memory requirements.  Thus, in our framework, we stick to the Compositional encoding for representing the Transition Relation. 

\begin{example}
% For each state in Fig \ref{fig: boolean_mani_game}, we assign boolean representation, for example, $v_{s1} = x_0 \wedge x_1 \wedge x_2$, $v_{s2} = x_0 \wedge x_1 \wedge \neg x_2$. Additionally, we represent each valid $a \in (A_s \cup A_e)$ with their corresponding boolean variables.
For each state and action in Fig \ref{fig: exp_no_human_mani_game}, we assign a corresponding boolean formula whose realization is set to true using variables $x \in X$, $i \in I$, and $o \in O$ as shown in Fig \ref{fig: boolean_mani_game}. Further, the transition relation $\eta$ from $v_{s1}$ and $v_{s4}$ is represented as follows.
% \noident
% From the initial state,
$\eta_{x_0}  = \eta_{x_2} = (x_0 \wedge x_1 \wedge x_2 \wedge o_0 \wedge o_1 \wedge \neg i_0).$
$\eta_{x_0}' = \eta_{x_1}' = \eta_{x_2}' =  (\neg x_0 \wedge x_1 \wedge \neg x_2 \wedge \neg o_0 \wedge \neg o_1 \wedge \neg i_0).$ 
% \begin{align*}
    % \label{eq: tr_init_state}
    % \eta_{x_0}' & = \eta_{x_1}' = \eta_{x_2}' =  (\neg x_0 \wedge x_1 \wedge \neg x_2 \wedge \neg o_0 \wedge \neg o_1 \wedge \neg i_0). \notag
    % \eta_{x_1} & =  (x_0 \wedge x_1 \wedge x_2 \wedge o_0 \wedge o_1 \wedge \neg i_0) \\
    % \eta_{x_2} & =  (x_0 \wedge x_1 \wedge x_2 \wedge o_0 \wedge o_1 \wedge \neg i_0) \notag
% \end{align*}
% Similarly for the state $v_{s4}$,
% \begin{align*}
    % \label{eq: tr_rstate}
    % \eta_{x_0}' = \eta_{x_1}' = \eta_{x_2}' =  (\neg x_0 \wedge x_1 \wedge \neg x_2 \wedge \neg o_0 \wedge \neg o_1 \wedge \neg i_0). \notag
    % \eta_{x_1}' & = (\neg x_0 \wedge x_1 \wedge \neg x_2 \wedge \neg o_0 \wedge \neg o_1 \wedge \neg i_0) \notag \\
    % \eta_{x_2}' & =  (\neg x_0 \wedge x_1 \wedge \neg x_2 \wedge \neg o_0 \wedge \neg o_1 \wedge \neg i_0) \notag
% \end{align*}
% \noident
% We take the union over each boolean variable $x \in X$ in Eq. \eqref{eq: tr_init_state} and Eq. \eqref{eq: tr_rstate}, i.e.,
Thus,  
$\eta_{x_0} = \eta_{x_0} \vee \eta_{x_0}'$, $\eta_{x_1} = \eta_{x_1} \vee \eta_{x_1}'$, $ \eta_{x_2} = \eta_{x_2} \vee \eta_{x_2}'$, and $\eta = \{\eta_{x_0}, \eta_{x_1}, \eta_{x_2}\}$. We repeat this for all edges in $\G_s$. 

% Further, we store the edges whose robot actions have the same cost as one monolithic TR. 
\end{example}

% As mentioned earlier, every valid edge on Symbolic Graph can be represented as boolean formula that evaluates to 1. This makes BDDs an ideal choice of representation. For graphs with edge weight, we use Algebraic Decision Diagrams (ADDs) \cite{bahar1997algebric} that allow us to represent the set of valid edges along with their corresponding edge weights. Each vaild path on the ADD graph point to a leaf node with integer value. Thus, BDDs can be thought as 0-1 ADD with leaf nodes restricted to 0 or 1.

% Additionally, using ADD allows us to perform boolean quantifier operations like existential and universal quantification, along with algebraic operations like $+$, $-$, $\cdot$, $\div $, $min$, and $max$  which are essential to perform Value Iteration over symbolic graphs.

% \subsection{Symbolic DFA}
% Formally we define it as follows.
\begin{definition}[Symbolic DFA $\mathcal{A}_{\phi}^{s}$]

Given DFA $\mathcal{A}_{\phi}$, its corresponding symbolic representation is defined as tuple $\mathcal{A}_{\phi}^{s} = (Y, X, \zeta, \mathcal{F})$, where
\begin{itemize}
    \item $X$ is as defined for $\G_s$ in Def.~\ref{def: comp_sym_game},
    \item $Y=\{y_0, y_1, \ldots, y_m\}$ is the set of boolean variables such that every state $z \in Z$ has corresponding boolean formula whose realization is true, % i.e., $2^{Y} \to 1$, 
    \item $\zeta = \{\zeta_{y_0}, \zeta_{y_1}, \ldots, \zeta_{y_m}\}$ 
    % \ml{still an infinite set}
    is a set of boolean functions, where $\zeta_{y_i}: 2^{X} \times 2^{Y} \to \{0,1\}$
    % , i.e., $\zeta(X, Y)$ evaluates to true iff $y_i \in Y$ is true in the successor state, and \km{$|\zeta| = |Y|$}, 
    % and $|\zeta| = |Y|$.
    such that $\zeta_{y_i}(X, Y) = 1$ 
    % \ml{$\zeta_{y_i}(X, Y) = 1$?}
    if $y_i \in Y$ is true in the successor state.
    % \ml{so, the evaluation of $\zeta_{y_i}$ does not depend on $X$?}\km{yes} \ml{why is it a function of $X$ then?}
    \item $\mathcal{F}$ is a boolean function whose realization is true for boolean formulas corresponding $z \in Z_f$.
    % \ml{make this finite as well!}
\end{itemize}
\label{def: sym_dfa}
\end{definition}

Given symbolic game $\G_s$ and DFA $\mathcal{A}^s_\phi$, we perform value iteration symbolically as described below.  
% Below, we describe our proposed procedure over symbolic graphs. 

\subsection{Symbolic Value Iteration (VI)}

% \km{add paragraphs for clarity.}

Here, we first show a VI algorithm for min-max reachability games with uniform edge weights.  Then, we extend the method to arbitrary edge weights. Finally, we show how this VI for reachability games can be adapted for DFA games, hence solving Problem~\ref{problem1}.

% \ml{add paragraph env.: VI for reachability games with uniform edge weights?}
% \noindent
\paragraph*{\textbf{VI with uniform edge weights}} 
In explicit VI in Alg.~\ref{algo: exp_value_iteration}, at every iteration of the loop, we need to iterate through every state $s\in S$ and apply the $\cPreMin$ and $\cPreMax$ operators (Lines 5-8). In fact, these inner loops are performed because we need to reason over a set of states at every (outer-loop) iteration. Indeed, this procedure can be naturally encoded in a symbolic fashion, where we can efficiently perform set-based operations. As VI is essentially a fixed-point computation, where we start from the accepting states and back-propagate the state values, we need to define the predecessor operation to (i) construct the set of predecessors, and (ii) reason over them.

We define the operator that computes the predecessors as the $\PreImage$ operator. 
% Alg.~\ref{algo: pre_mono} outline this predecessor computation using ADDs. 
% \ml{can we have a pseudocode for it?}
% Say, we have a set of target states.
Let $\omega$ be the boolean function representing a set of target
% \ml{hmm.. what's a traget states in $\G$?}
states $\mathcal{F}'$ in $\G_s$, i.e., $\omega(X): 2^X \to 1$
% evaluates to true 
for states in the target set $\mathcal{F}'$. 
% Here $X$ is the set of boolean variables as defined in Definition \ref{def: comp_sym_game}. 
Now, to compute the predecessors, we use a technique called \textit{variable substitution}. For each boolean variable $x \in X$ in $\omega$, we substitute the variable $x_i$ with its corresponding transition relation $\omega(\eta_{x_i})$. Variable substitution using ADDs can be implemented using the $\Compose$ operator shown on Line \ref{algo: pre_mono-compose} in Alg. \ref{algo: pre_mono}.
Substituting $x_i \leftarrow \eta_{x_i}$ in $\omega$ captures the value of $x_i$ in the predecessor states for all valid actions in $\eta_{x_i}$.  We
% To construct the set of valid predecessors, 
repeat the process for all $x_i \in X$ and store it as $pre$. Thus,
% $\omega_{k+1}$.
% Note,  $k$ corresponds to the $k^{th}$ application of the $\PreImage$ operator while
$pre(X, I, O)$ is the new boolean function
% defined over $X$, $I$, and $O$ where
such that $2^X \times 2^I \times 2^O \to 1$ for the set of valid edges from states in $pre$ to a state in $\omega$. 
% Line \ref{algo: pre_mono-compose} in Alg. \ref{algo: pre_mono} 
% % \ml{need to refer to this algorithm earlier...}
% shows how to perform this operation using the $\Compose$ operator for ADDs
% \ml{what's a $\Compose$ operator?}
% .
% \ml{the last sentence is confusing!}
% If we apply $\PreImage$ operation continuously over $\omega$, we will reach a fixed point, i.e., $\omega_{k+1} \equiv \omega_{k}$($\equiv$ is logical equivalence), where $\omega_{k+1}$ is the set of all reachable states when playing a min-min (cooperative game). 

% Similarly,
To compute the set of winning states for a reachability game with no quantities,
we perform the following procedures. 
First, we perform universal quantification. This preserves all the states that can \textit{force} a visit to the next state. Then we  existentially quantify the boolean function to compute set of states that can force a visit in one-time step. Mathematically,  $\omega_{1} = \exists O \cdot ( \forall I \cdot pre(X, I, O))$. 
% After every iteration of $\PreImage$, we perform universal quantification under all human actions, i.e., $\forall I \cdot pre(X, I, O)$. This preserves all the states that can \textit{force}  a visit to a state in $\omega$ from a state in $pre$.
% $2^{I} \to 1$, 
% then we can compute the set of states from which the robot can \textit{force} a visit to a state in $\omega$ from a state in $pre$.
% Intuitively, the universal operation translates to - check if a transition exists from states $\omega_{k+1}$ under all human actions to states in $\omega_{k}$. 
% We then existentially quantify the boolean function, i.e., $\omega_1 = \exists O \cdot pre(X, O)$
% $\omega_{k+1}$ 
% to compute the set of reachable states that can force a visit to the target states in one step.
We repeat this process till we reach a fixed point, i.e., $\omega_{k-1} \equiv \omega_{k}$.

To compute the winning strategies, we initialize an additional boolean function $t$ as $t(X, O)_0 = \omega_0$.
% as the strategy for the Sys player is to choose any valid action after the task has been completed. 
After each fixed-point computation, $t(X,O)_{k+1} = \forall I \cdot pre(X, I, O)_{k}$. 
% we can universally quantify $\omega$ to enforce that an edge exists under all possible human actions. 
For quantitative reachability games, we replace the universal operation with $\max$ and the existential operation with $\min$. Thus, for a graph with uniform edge weights, this method suffices. 
Next, we discuss how to perform VI over $\G_s$ with arbitrary edge weights.
% \ml{$\G_s$ has no quantity}.
% \ml{some where above, we need to say that we first discuss when edge weights are equal, and then value iteration over $\G_s$}

% For a graph with arbitrary edge weights, we 
% slightly modify the algorithm.

\paragraph*{\textbf{VI with arbitrary edge weights}} 
% Alg. \ref{algo: sym_value_iteration} outlines the pseudocode for this section.
Similar to the case of the uniform weights, we start with $\omega_0$. But instead of storing it in a monolithic boolean function, we decompose it into smaller boolean functions, each corresponding to the value associated with the states.  Mathematically, $\omega_k = \langle B_j \rangle$,  where $\langle B_j \rangle$ is a sequence of boolean functions, each representing the set of states with value $j$. 
For the initial iteration, $k = 0$,  $j = 0$, and $B_j \equiv \omega_0$ is the boolean function for $\mathcal{F}'$. To compute the predecessors, we first separate the transition relation $\eta$ according to its  edge weights, i.e., $\eta = \{\eta_{c_1}, \eta_{c_2}, \ldots, \eta_{c_n} \}$, 
where $c_i$ is edge weight associated with edges in $\G$.
% \km{and $|\eta|$ is finite as defined below.}
% and $\eta_{w_1}$.
% We define this as the Monolithc Representation and is formally defined in Definition $\ref{def: mono_tr}$.
% as follow
% We can segregate the transition relation based on their corresponding edge weights or for each robot action $A_s$. We call the former Monolithic representation and the latter Partitioned representation. Formally,

\begin{algorithm}[tb!]
    \caption{$\PreImage$ 
    % \ml{$\PreImage(..., \G_s)?$}
    }
    \label{algo: pre_mono}
    \SetKwInOut{Input}{Input}\SetKwInOut{Output}{Output}
    
    \SetKwData{sval}{sval}
    \SetKwData{Win}{Win}
    \SetKwData{aPre}{aPre}
    \SetKwData{uPre}{uPre}

    \SetKwData{ADD}{ADD}

    \SetKwData{Strategy}{Strategy}
    \SetKwData{hStrategy}{hStrategy}

    %  ADD and BDD operations
    \SetKwFunction{addZero}{addZero}
    \SetKwFunction{addInfinity}{addInfinity}
    \SetKwFunction{restrict}{restrict}
    \SetKwFunction{min}{min}
    \SetKwFunction{max}{max}
    \SetKwFunction{add}{add}
    \SetKwFunction{ite}{ite}
    \SetKwFunction{preimage}{preimage}
    \SetKwFunction{vectorCompose}{vectorCompose}
    
    \SetKwData{This}{this}\SetKwData{Up}{up}
    \SetKwFunction{ValueIteration}{ValueIteration}

    \Input{vector $\langle \ADD(w_k) \rangle$, Symbolic Game $\G_s$, 
    % state values $\sval$,
    }
    \Output{
    % $\sval$,
    vector of pre-images $\ADD(pre)$}
    $pre = \langle \emptyset \rangle$ \hspace{6mm} \small{// vector to store predecessor as 0-1 ADDs} \\
    \For{$\ADD(B_j) \in \ADD(w_k)$}{
        \For{$\eta_{c_i} \in \eta$}{
            % $ract \leftarrow \text{get weight associated with TR} \; \eta_i$ \\
            $j' = j + c_i$ \\
            % $\sval.\add(j')$ \\
            
            $\ADD(B_{j'}) = \ADD(B_j).\vectorCompose(X, \eta_{c_i})$ \\ 
            \label{algo: pre_mono-compose}
            %\tcc*[r]{0-1 ADD}
            % $\ADD(pre)_c = \preimage(\ADD(w))$ \tcc*[r]{0-1 ADD}   
            $pre_{j'} = \ADD(B_{j'})$ \\ %\tcc*[r]{store it in vector} 
        }
    }
    \KwRet{
    % $\sval$,
    $\ADD(pre)$}
\end{algorithm}

\begin{definition}[Monolithic Transition Relation]

A Monolithic Transition Relation for $\G_s$ is defined as $\eta = \{\eta_{c_i}(X, I, O)\}$ 
% \ml{what is $i$?}
where $\eta_{c_i} = \{ \eta_x | x \in X \}$
% \ml{what is $i$?}
% is a sequence of boolean functions,
and $c_i$ is an edge weight in the set of edge weights given by $F$. $\eta_{c_i}$ is a set of boolean functions and $\eta$ is separated based on the set of all edge weights in $\G$, i.e., $|\eta| = |\{F(v, a_s)\}|$ $\forall v \in V$ and $\forall a_s \in A_s$. Further, an edge appears only once in $\eta$, i.e., $\eta_{c_i}(\eta_{x_i}) \cap  \eta_{c_j}(\eta_{x_i}) = \emptyset$ for every $i \neq j$. 
\label{def: mono_tr}
\end{definition}

\begin{definition}[Partitioned Transition Relation]
% \ml{same comments here}
A Partitioned Transition Relation for $\G_s$ is defined as $\eta = \{\eta_{c_i}(X, I, O)\}$ where $\eta_{c_i} = \{ \eta_x | x \in X \}$ is a set of boolean functions.
% each defined over $X, I, O$.
Each $\eta$ is separated by robot actions $a_s \in A_s$ i.e., $|\eta| = |A_s|$ and $c_i \in \{F(v, a_s)\}$ $\forall v \in V$ and $\forall a_s \in A_s$.
\label{def: part_tr}
\end{definition}
% When $\eta$ is segregated based on actions in $A_s$, we call it Partitioned Transition Relation. Formally,

Thus, for each $\eta_{c_i}$ and for each $B_j \in \omega_k$, we compute the $\PreImage$ of $B_j$ and store 
them as $\omega_{k+1} = \langle B_{j'} \rangle$ where $j' = j + c_i$. If  $B_{j'}$ 
already exists, we take the union of the boolean functions. Alg.~\ref{algo: pre_mono} summarizes the pre-image computation over weighted edges.
% \ml{I thought Alg. 2 was for uniform edge case!}
% We repeat this process until $\omega_{k+1} \equiv \omega_{k}$, i.e., every boolean function $\B_{j}$ for $k+1$th iteration is $\equiv B_{j}$ in $k$th iteration
% \ml{a little confused... which algorithm are you explaining here?} 
% Intuitively, at every iteration, we add the states with non-infinity state value to the set of reachable states. 
% After reaching the fixed point, we have computed the optimal value from all states, and we simply return w.
For strategy synthesis, we repeat the same process as above until $\omega_{k+1} \equiv \omega_{k}$,
i.e., every boolean function $B_{j}$ in the $(k+1)$th iteration is $\equiv B_{j}$ in the $k$th iteration.
% For pseudocode, see the extended version \cite{extended_paper}. 
Alg.~\ref{algo: sym_value_iteration} gives the pseudocode for symbolic VI with quantitative constraints. For efficient manipulation of $B_{j}$, we use ADDs. We use the $\Compose$ operation for substitution,
and $\min$ and $\max$ for algebraic computations. 

To check for convergence, we exploit the canonical nature of ADDs for constant time equivalence checking.

% \noindent
% \paragraph*{Algorithm \ref{algo: pre_mono}} 
% \ml{could remove this paragraph.}
% We initialize the algorithm with the ADD representation of $\omega_0$ corresponding to set of target states $\mathcal{F}'$. During every fixpoint computation, we first convert ADD($\omega_k$) to a vector of 0-1 ADDs using the $\bddInterval$ operator. We then compute the predecessors as per Algorithm \ref{algo: pre_mono}. We then iterate over the vector of 0-1 ADDs and conjoin them to construct a monolithic ADD. Finally, we perform $\max$ and $\min$ for min-max computation. We repeat this process until ADD($\omega_{k+1}$) $\equiv$ ADD($\omega_{k}$). Note we skip the equivalence checking for the 1st iteration. 

\begin{algorithm}[t!]
    \caption{Symbolic Value Iteration}
    \label{algo: sym_value_iteration}
    \SetKwInOut{Input}{Input}\SetKwInOut{Output}{Output}
    \SetKwData{sval}{sval}
    \SetKwData{Win}{Win}
    \SetKwData{aPre}{aPre}
    \SetKwData{uPre}{uPre}

    \SetKwData{ADD}{ADD}

    \SetKwData{Strategy}{Strategy}
    \SetKwData{hStrategy}{hStrategy}

    %  ADD and BDD operations
    \SetKwFunction{set}{set}
    \SetKwFunction{addZero}{addZero}
    \SetKwFunction{addInfinity}{addInfinity}
    \SetKwFunction{restrict}{restrict}
    \SetKwFunction{min}{min}
    \SetKwFunction{max}{max}
    \SetKwFunction{argmax}{argmax}
    \SetKwFunction{argmin}{argmin} 
    \SetKwFunction{add}{add}
    \SetKwFunction{ite}{ite}
    \SetKwFunction{preImage}{preImage}
    \SetKwData{This}{this}\SetKwData{Up}{up}
    \SetKwFunction{ValueIteration}{ValueIteration}   
    \Input{Symbolic Two-player Game $G_s$}
    \Output{Winning strategy $\ADD(\sigma)$, Optimal values $\ADD(\omega_k)$}
    % initial set of winning state to be zero
    % $\sval = \set\{0\}$ \\ %\tcc*[r]{set of state values}
    % $\ADD(\omega_0) \leftarrow $ Initialize 0-$\infty$ $\ADD$ \\
    $\ADD(\sigma) \leftarrow $ Initialize  $\infty$ ADD \\
    $\ADD(\tau) \leftarrow $ Initialize  $0$ ADD \\
    % $\ADD(\omega^{'}_0) \leftarrow $ Initialize  $0$ ADD \\
    $\ADD(\omega_0), \ADD(\mathcal{F}') \leftarrow$ \text{\small{Init 0-$\infty$ $\ADD$ for target states}} \\
    $\ADD(\omega_0) = \ADD(\omega_0).\min(\ADD(\mathcal{F'}))$\\
    $\ADD(\sigma) = \ADD(\sigma).\min(\ADD(\omega_0))$  \\
    \While{$\ADD(\omega_{k-1}) \not \equiv \ADD(\omega_k)$}{
        % $\ADD(\omega'_k) = \ADD(\omega_k) $ \\
        $\langle \ADD(B_{j}) \rangle \leftarrow \text{\small{Construct 0-1 $\ADD$s from $\ADD(\omega_k)$}}$ \\ %where $j \in \sval$}$ \\
        % $\sval,
        $\ADD(pre) \leftarrow \preImage(\langle \ADD(B_j) \rangle, G_s)$ \\
        % , \sval)$ \\

        % \tcc{get all predecessors}
        $\ADD(\aPre) \leftarrow $ Initialize  $0$ ADD \\
        \For{$\ADD(B_j) \in \ADD(pre)$}{
            $\ADD(\aPre) = \ADD(\aPre).\max(\ADD(B_j))$
        }
        \For{$f \in 2^{I} \to 1$}{
            $\ADD(\tau) =   \ADD(\tau).\max(\ADD(\aPre).\restrict(f))$\\
        }
        
        $\ADD(\sigma) =   \ADD(\sigma).\min(\ADD(\tau))$ \\
        \For{$f \in 2^O \to 1$}{
            $\ADD(\omega_{k + 1}) =   \ADD(\omega_k).\min(\ADD(\sigma).\restrict(f))$\\
        }
    } 
    \KwRet{$\ADD(\omega_k)$, $\ADD(\sigma)$}
\end{algorithm}

\subsection{Strategy Synthesis for symbolic DFA Games}

To synthesize symbolic winning states and strategies for the DFA game $\PA$, we compose
the $\PreImage$ computation as described above
to evolve over $\G_s$ and $\mathcal{A}_\phi^s$ in an asynchronous fashion. Given a boolean function representing the set of accepting states such that $\omega_0(X, Y) = 2^{X} \times 2^{Y} \to 1$ ($2^X \to 1 \quad \forall v \in V$ and $2^Y \to 1 \quad \forall z \in Z_f$), we first 
compute the predecessors over the states of $\mathcal{A}^{s}_{\phi}$ by evolving over $y \in Y$,
i.e., $pre_{\phi}(X, Y) \leftarrow  \PreImage$ of $\omega_0(X, Y)$. This intermediate boolean function captures only the evolution of the DFA variable, i.e., $2^{X} \times 2^{Y} \to 1$ 
such that $2^{Y} \to 1$ for the valid DFA states in the predecessors. Finally, we evolve over the symbolic game $\G_s$
from the intermediate boolean function, i.e., $pre(X, Y, I, O) \leftarrow \PreImage$ of $pre_{\phi}(X, Y)$, to compute the boolean function representing valid predecessors in $\G_s$. Algorithmically, we modify Alg.~\ref{algo: pre_mono} with the intermediate $pre_{\phi}$ computation and keep the $pre$ computation consistent for evolution over $pre_{\phi}$.

We employ the above algorithm to efficiently synthesize symbolic winning strategies for Problem \ref{problem1}. 
Our evaluations illustrate that this symbolic algorithm has an order of magnitude speedup over the explicit approach. For Problem~\ref{problem2}, previous work \cite{muvvala2022regret, filiot2010iterated} provide algorithms for synthesizing regret-minimizing strategies for explicit graphs. Next, we discuss our efficient symbolic approach to this problem and synthesis of regret-minimizing strategies.

\section{Hybrid Regret synthesis algorithm }

The first step in synthesizing regret-minimizing strategies is to construct the Graph of Utility $\G^u$ \cite{muvvala2022regret}. Intuitively, this graph captures all possible plays (unrolling) over $\PA$ along with the total payoff associated with each path to reach a state in $\PA$. In the explicit approach, we first construct the set of payoffs as $[B] = \{0, 1, 2, \ldots, \B \}$ where $\B$ is the user-defined budget. We then take the Cartesian product $\PA \times [B]$ to construct states $(s, u)$ of $\G^u$. An edge $(s, u) \to (s', u')$ in $\G^u$ exists iff $(s, a, s')$ is a valid edge in $\PA$ and $u' = u + F(s, a)$. As constructing $\G^{u}$ is Pseduo-polynomial, this computation is memory intensive.

% ]For more details refer to \cite{muvvala2021thesis}. \ml{say this is memory intensive}

\begin{figure*}[t]
    \centering
    \begin{subfigure}[t]{0.3\textwidth}
        \centering
        % \hspace{-3mm}
        % \includegraphics[width=1.02\linewidth]{images/adv-results/scenario2/loc_synth_time.png}
        \includegraphics[width=.99\linewidth,valign=t]{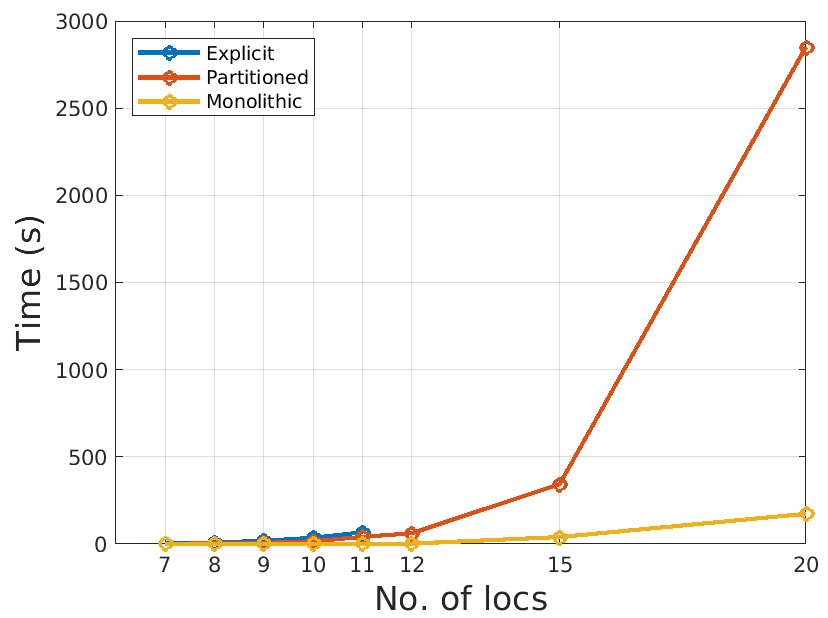}
        \caption{\small $|O| = 3$ and variable $|L|$}
    \label{fig: adv-scenario2}
    \end{subfigure}%
    ~~
    \begin{subfigure}[t]{0.3\textwidth}
        \centering
        % \hspace{-5mm}
         % \includegraphics[width=0.99\linewidth]{images/adv-results/scenario3/box_synth_time.png}
         \includegraphics[width=0.99\linewidth,valign=t]{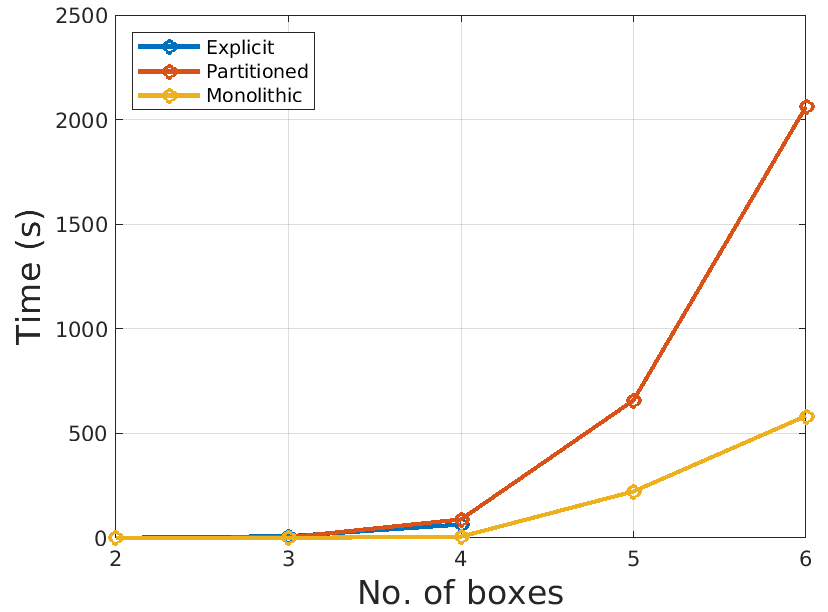}
        \caption{\small $|L| = 8$ and variable $|O|$}
    \label{fig: adv-scenario3}
    \end{subfigure}%
    ~~
    \begin{subfigure}[t]{0.3\textwidth}
        \centering
        % \vspace{2mm}
         % \includegraphics[width=0.99\linewidth]{images/adv-results/scenario3/box_synth_time.png}
         \includegraphics[width=0.97\linewidth, valign=t]{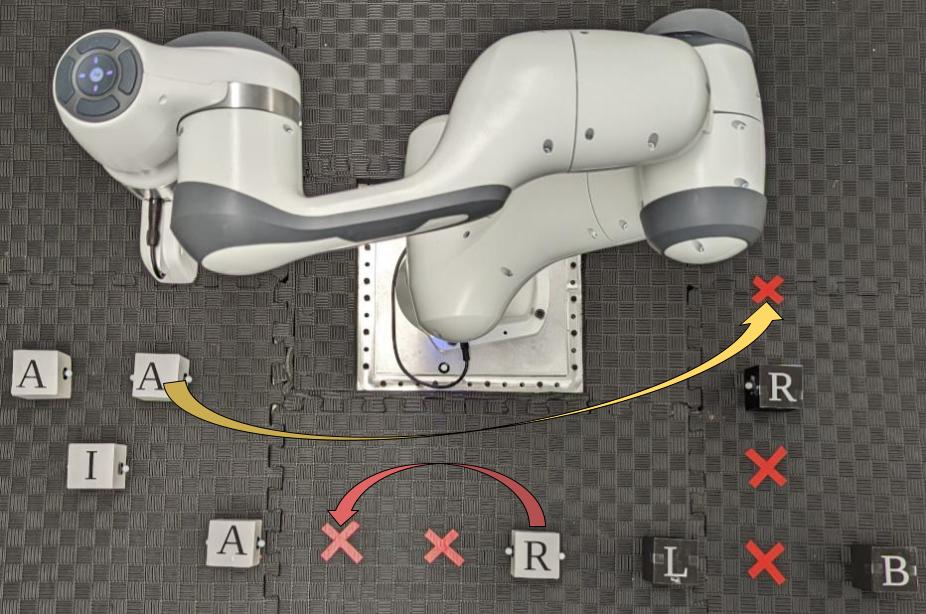}
         \vspace{5mm}
        \caption{\small Regret-minimizing strategy}
    \label{fig: regret-game1}
    \end{subfigure}%
\caption{\small (a) and (b) Benchmark results for min-max synthesis. (c) Regret-minimizing strategy for the task in Fig.~\ref{fig: strs_illustration}.}
\label{fig: experiemt_min_max}
\vspace{-1.mm}
\end{figure*}

\begin{figure*}[ht]
    \centering
    \begin{subfigure}[t]{0.3\textwidth}
        \centering
        \includegraphics[width=0.99\linewidth]{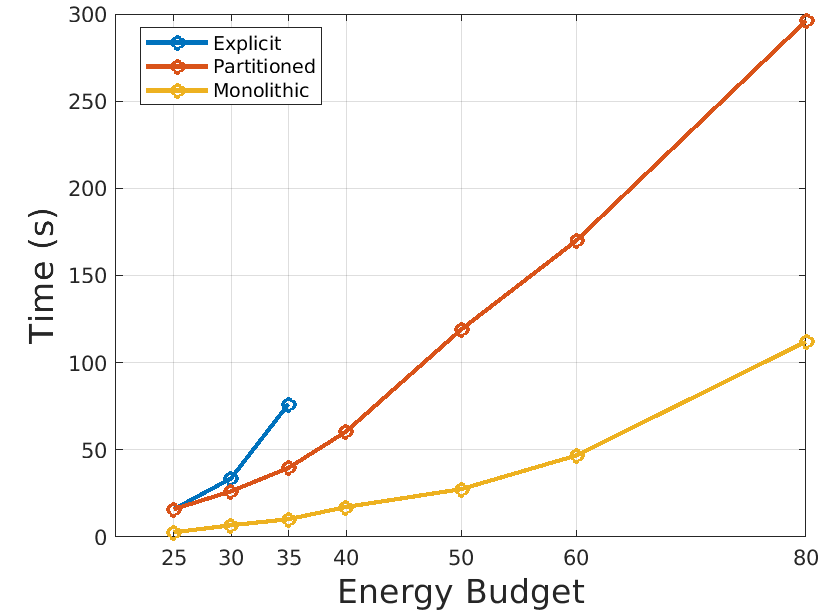}
        \caption{\small $|O| = 3$, $|L| = 7$, variable $\B$}
    \label{fig: reg-scenario1}
    \end{subfigure}%
    ~~
    \begin{subfigure}[t]{0.3\textwidth}
        \centering
         \includegraphics[width=0.99\linewidth]{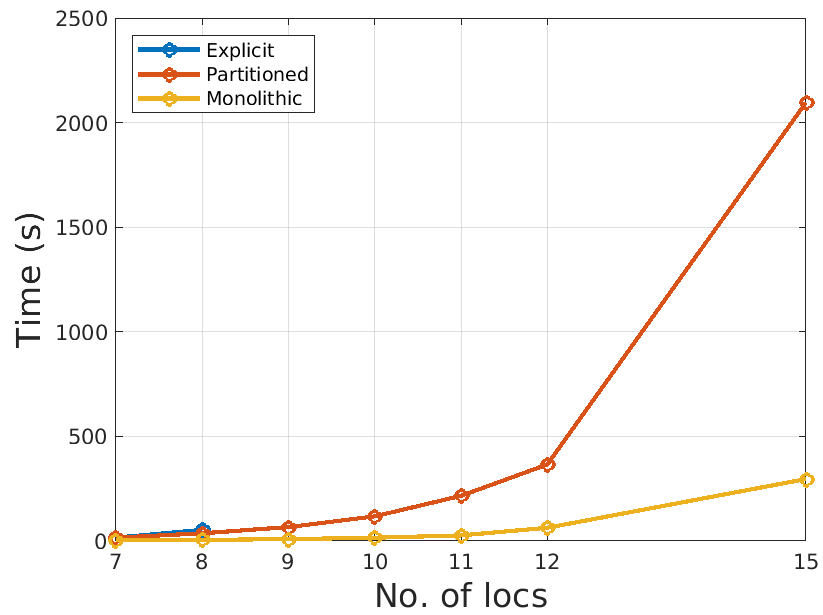}
        \caption{\small $|O| = 3$, $\B = 25$, variable $|L|$}
    \label{fig: reg-scenario2}
    \end{subfigure}%
    ~~
    \begin{subfigure}[t]{0.3\textwidth}
        \centering
         \includegraphics[width=0.99\linewidth]{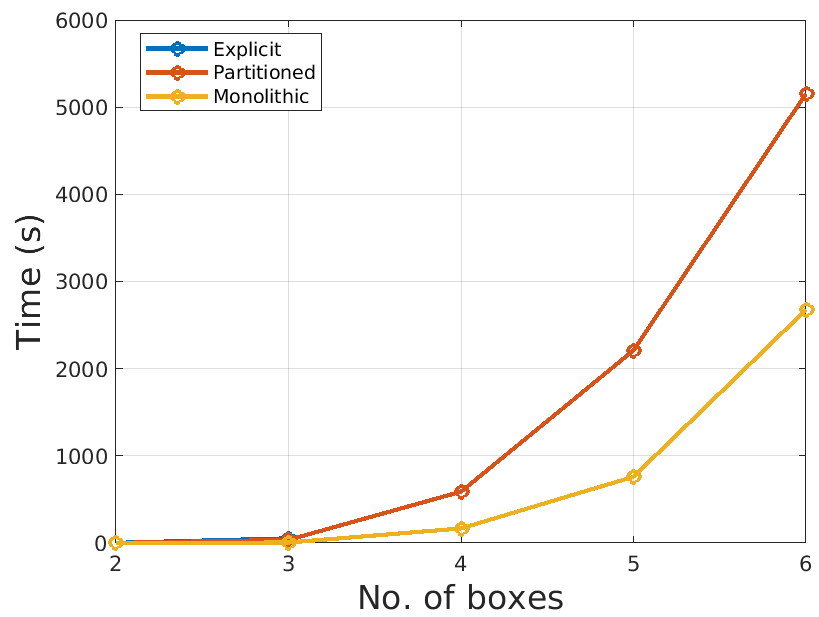}
        \caption{\small $|L| = 8$, variable $\B$ \& $|O|$}
    \label{fig: reg-scenario3}
    \end{subfigure}%
\caption{\small Benchmark results for regret-minimizing strategy synthesis in Problem \ref{problem2}}
\label{fig: experiemt_reg}
% \vspace{-4.5mm}
\end{figure*}

To construct $\G^u$ efficiently, we define $\G^u_s$ as the symbolic Graph of Utility. To construct the transition relation, we create an additional set of boolean variables $u \in U$ such that $2^U \to 1$ for valid utilities in $[B]$. We then create the transition relation $\eta_{\, \text{\scriptsize U}}$ for the utility variables to model their evolution symbolically under each action $A_s$. Mathematically, $\eta_{\, \text{\scriptsize U}} = \{ \eta_{u_1},  \eta_{u_2}, \ldots, \eta_{u_n} \}$, where $\eta_{u_i} = \{\eta_u | u \in U\} \forall i \in n$ and $|\eta_{\, \text{\scriptsize U}}| = |\eta|$. Thus, to perform $\PreImage$ operation over $\G^u_s$, we substitute variables $x_i \in X$ and $u_i \in U$ in a synchronous fashion using the $\Compose$ operator.
% \ml{again $\Compose$!}

% Note that ``unrolling" a graph is a crucial step in many games-played-on-graphs and hence our approach to symbolically unrolling a graph can be utilized in other frameworks.

The second step in regret-based synthesis is to construct the graph of the \textit{best response} $\G^{br}$. This graph captures the best alternate payoff $ba$ from every edge along each path on $\G^u$, i.e., $\min_{\sigma'} \Val(\sigma',\tau)$ in \eqref{eq: reg_def}. For a fixed strategy
$\sigma$, we compute $ba =  \min_{\sigma' \in \Sigma^{\G} \backslash \sigma }\, \min_{\tau \in \Tau^{\G}}\, \Val^{(s, u)}(\sigma, \tau)$ assuming human to be cooperative (see \cite{filiot2010iterated} for formal definition).
% and use Alg.~\ref{algo: sym_value_iteration} for symbolic computation. 
% $ba = \cVal^{(s, u)}(\G^u)$ where $\cVal$ is the payoff associated with states $(s, u)$ under the cooperative human assumption
% for all actions from $(s, u)$ except $\sigma \in \Sigma^\G$. 
% We can compute $\cVal$ using Alg.~\ref{algo: sym_value_iteration} by taking $\min$ over both human and robot actions for $\G_s^u$.
Finally, we repeat the process for every edge in $\G^u_s$ to construct the set of $ba$.
% loop over each edge in $\G^u_s$ and compute $ba$.

For symbolic implementation, we first compute cooperative values by first playing min-min reachability game
% $\cVal$ over $\G^u_s$ using Alg.~\ref{algo: sym_value_iteration}
and then ``explicitly" loop over every edge in $\G^u_s$ to construct the symbolic transition relation for the symbolic graph of best response $\G^{br}_s$. While the computation of the $ba$ set is explicit, the resulting transition relation for $\G^{br}_s$ is purely symbolic. Finally, we play a min-max reachability game
using the method described above
% Alg.~\ref{algo: sym_value_iteration}
to compute regret-minimizing strategies over $\G^{br}_s$. Our empirical results show that using this approach scales to
larger manipulation problems with significantly lower memory consumption than the explicit approach.
% twice as many objects and thrice as many locations as the explicit approach with significantly lower memory consumption.
% This algorithm requires the construction of two additional graphs - Graph of Utility and a Graph of Best Response. We will talk discuss how to construct these graphs symbolically. Also, mention that the graph of bets response is hybrid in nature as it explicitly loops over all edges in the graph of utility and constructs the edges symbolically.   

\section{Experiments}
\label{sec: results}

We consider various scenarios for our pick-and-place manipulation domain. We benchmark the explicit approach with the symbolic approaches from Section \ref{sec: approach} for min-max reachability games in Problem \ref{problem1} and regret-minimizing strategy synthesis for Problem \ref{problem2} with varying input parameters, i.e., energy budget $\B$, number of locations $|L|$, and number of boxes $|O|$. 
% For problem \ref{problem1}, we benchmark computation time for synthesizing winning strategies
% using the Explicit, Monolithic, and Partitioned Symbolic approaches. 
% and 
% for Problem, \ref{problem2}, we benchmark computation times for regret synthesis. Specifically, we consider three scenarios. In Scenario one, we fix the number of objects $|O|$ and the number of locations $|L|$ and vary the budget $\B$. For the second scenario, we fix $|O|$ and $\B$ and increase $|L|$. Lastly, for scenario 3, we fix $|L|$ and vary $\B$ and $|O|$.
% vary the budget and the number of objects in the workspace. 
The task for the robot in all scenarios is to place boxes in their desired locations. 
% For Scenario
% box 0 or box 1 in their desired locations as shown in Fig. \ref{fig: experiemt_setup}. The task is $\phi = F(p_{00} | p_{17})$.
% Note, for Scenario 2 and Scenario 3, the desired locations are blocked thus the robot need

For these benchmarks, we consider the same experiments as in \cite{muvvala2022regret}, where
the robot spends 3 units of energy for every action to operate away from the human and one unit of energy when operating near the human. Refer to \cite{muvvala2022regret} 
% \ml{and the extended version.}
for the experimental setup details for Problem \ref{problem2}. 
% benchmarks and the comparison between regret-minimizing vs. min-max strategies.
% performed within the Robot Region otherwise and one unit of energy in the human region.
% and ``transit" and ``transfer" to the Robot Region.
% The robot spends one unit of energy to operate purely in the human region.
% We construct the abstraction by first parsing the PDDL files and then constructing the Symbolic and the explicit graph as outlined in \ref{sec: approach}.
% All the experiments are run single-threaded on an Intel i5 -13th Gen 3.5 GHz CPU with 32 GB RAM. 
% \ml{to save space, the following is not needed. just keep the github page.}
% \ml{mention demo in fig1 that explicit did not work}
% \footnote{GitHub: https://github.com/MuvvalaKaran/symbolic\_synthesis}
% We use MONA for the translation of LTLf to DFA \cite{henriksen1995mona}.
Then, we demonstrate our symbolic synthesis algorithm on the 5-box scenario in Fig.~\ref{fig: strs_illustration}, on which the explicit approach runs out of memory.
% we could not synthesize a strategy for this scenario using the explicit approach.
% and employ PyPerplan to parse the PDDL files for constructing all the propositions $\Pi$ in $\G$. 
% related to the Task
% and Robot configurations. We then
% and assign a unique boolean representation to each proposition. 
% Note, we use variables $X_i \subset X$ where $i$ is equal to the number of boxes plus 1 (for robot conf.) to uniquely represent each state in $G$ such that $X_i \cap X_j = \emptyset$ for $i \neq j$. Finally, a set of states in the graph can be symbolically represented as an ADD associated with the Boolean function $f(X) = \bigvee \left( \bigwedge X_i \right)$ where the realization of $\bigwedge X_i$ is true for $s \in G$.
% \qh{Should discuss demonstration as well - "Finally, we demonstrate our symbolic synthesis algorithm on a ..."}

Our implementation is in Python. It uses a Cython-based wrapper for the CUDD package for ADD operations \cite{somenzi2020cudd}, and MONA for LTLf to DFA translation~\cite{henriksen1995mona}. The tool is available on GitHub \cite{repolink}.

\textbf{Min-Max with Quantitative Constraints: }
For problem~\ref{problem1}, we benchmark computation time for synthesizing winning strategies using the Explicit, Monolithic, and Partitioned symbolic approaches. For scenario 1, we fix $|O| = 3$ and vary $|L|$. In scenario 2, we fix $|L| = 8$ and vary $|O|$.
 % The computation times for fixed $|O| = 3$ and variable $|L|$ is shown in Fig. \ref{fig: adv-scenario2}
 % . For scenario 1, the explicit approach ran out of memory at $|L| = 11$,
 % while the symbolic approaches could scale to 2x more locations with reasonable computation times.
 Results are shown in Fig.~\ref{fig: experiemt_min_max}.
 For the Monolithic approach, we see at least an order of magnitude speed-up over the Explicit approach for both scenarios. Further, there is almost an order of magnitude speedup for the Monolithic approach over the Partitioned approach for scenario 1. 
 % Similar results are observed in \cite{he2019symbolic}. 
 We note that the Partitioned approach is slower than the Monolithic approach due to the size of the Monolithic ADD not being large enough to compensate for the computational gain from the $\PreImage$ operation over multiple smaller ADDs. Hence, the cumulative time spent computing predecessors is much more in the Partitioned approach for every fixpoint iteration than in the Monolithic approach. For the Explicit approach, we could not scale beyond 4 boxes and 8 locations due to the memory required to store $\G_s$ explicitly ($>$ 15 GB). 
 % which slows down the Partitioned approach.
 % \ml{comments on explicit}

 % for the larger instances $|L| \geq 15$. The comp
\textbf{Regret-Minimizing Strategy Synthesis: }
% We benchmark our framework for three Scenarios and note their computation times. We note that for the explicit approach, we were not able to scale beyond 3 boxes and 7 locations, while the symbolic approach was able to synthesize strategies for 5 objects and 10 locations with reasonable computation times and significant memory savings. 
For problem \ref{problem2}, we consider three scenarios. In scenario 1, we fix $|O|$ and $|L|$ and vary $\B$. For the scenario 2, we fix $|O|$ and $\B$ and increase $|L|$. Finally, for scenario 3, we fix $|L|$ and vary $\B$ and $|O|$. 
Note, for scenario 3, the regret budget $\B$ is chosen to be 1.25 times the minimum budget from Problem \ref{problem1}. 
Results are shown in Fig.~\ref{fig: experiemt_reg}.
For all the scenarios, we observe that the explicit approach runs out of memory for relatively smaller instances. Additionally, we observe that overall the Monolithic approach is faster than the Partitioned approach due to the size of the Transition relation, as mentioned above. 

% For the symbolic implementation, we compare the Monolithic Transition Relation representation with the Partitioned Transition Relation. We find that using Monolithic Transition Relation gives us at least $5 \times$ speed up over the partitioned implementation. We believe this is due to the number of the ``compose"  function call. All computation results are averaged over 5 runs. 

\textit{Scalability on the Budget: } Unrolling the explicit graph is a computationally intensive procedure as the state space grows linearly with budget $\B$. Thus, we observe in Fig. \ref{fig: reg-scenario1} that the explicit approach runs out of memory for $\B > 35$. There are $62,800$ explicit nodes for $\B = 35$ and $295,496$ explicit nodes for $\B = 80$ in $\G^{br}_s$. We required maximum of 860 MB for the symbolic approaches and more than 9 GB for the explicit approach with $\B = 35$.
% For the symbolic approaches, the maximum memory required for ADDs is 860 MB \& 9 GB of memory is needed to store the graphs explicitly for $\B = 35$.
% \ml{how about explicit?}

\textit{Scalability on the number of locations: } For this scenario, we observe that the Monolith approach is at least 3 times faster than the Partitioned approach. This is because, as we increase the number of locations, we do not necessarily need more boolean variables. The gap widens as we keep increasing $|L|$. For $|L| = 20$, the graph has $~7 \times 10^{6}$ nodes and $55 \times 10^{6}$ edges in $\G^{br}_s$. The maximum memory consumption is $1040$ MB as opposed to over $5$ GB memory for $|L| = 8$. 

\textit{Scalability on the number of objects: } For scenario 3, we observe that the explicit approach does not scale beyond 3 objects. The computation times for the symbolic approach also increase exponentially due to the fact that every object added to the problem instance needs an additional set of boolean variables.  For $|O| = \{2, 3, 4, 5, 6\}$, we needed 29, 36, 40, 45, and 49 boolean variables with a maximum of 1100 MB consumed. The explicit approach required above 15 GB of memory for $|O| = 4$.  

\textbf{Physical Execution:} We demonstrate the regret-minimizing strategy for building  ``ARIA LAB" in Fig.~\ref{fig: strs_illustration}. Figs. \ref{fig: regret-game1} and \ref{fig: strs_illustration} show the initial and final configuration for $|O| = 5$ (white boxes) and $|L|=7$. Robot and human cannot manipulate black boxes. We note that the human can only move the boxes in its region to its left. The robot initially operates away from the human in the robot region (yellow arrow in Fig. \ref{fig: regret-game1}). The human moves ``R" (red arrow)  cooperatively and allows the robot finish the task in its region (Fig. \ref{fig: reg_beh}).

% The ifigure shows the initial conf. of boxes and the task is to build ``ARIA LAB" as shown in Fig. \ref{fig: strs_illustration}

% The number of boolean variables needed for every object and the added memory requirements is shown in Table \ref{table: boxes_bVars}. 

% \subsection{Discussion}
\textbf{Discussion: }
We show through our empirical evaluations that symbolic approaches can successfully scale to larger problem instances with significant memory savings. We observe that using a Monolithic representation of the transition relation is faster than the Partitioned approach due to the size of ADDs. Further, while this paper focuses on boolean representation for our problem, other ADD based optimizations like variable ordering can be utilized for a more compact representation of ADDs which could lead to additional computational speedups. Finally, we note that we compute the set of reachable states on $\G^u_s$ before VI. This allows the synthesis algorithm to only reason over the set of reachable states and thus speed up the synthesis process. 
% \km{In future, we plan to explore non-zero sum games or alternate notions of strategies to relax the adversarial assumption to model a strategic human with its own objective.}

% \begin{table}[t!]
% \caption{Number of boolean variables and memory needed for Problem \ref{problem2} Scenario3.
% % Note the increase in memory to store ADDs.
% }
% \centering
% \begin{tabular}{c || c| c}
% \toprule
% Boxes $\left( |O| \right)$ & boolean variables & Memory (MB) \\
% \hline \hline
% 2 & 29 & 73.776 \\
% 3 & 36 & 723.754 \\
% 4 & 40 & 971.132 \\
% 5 & 45 & 1016.007 \\
% 6 & 49 & 1067.688 \\
% \bottomrule
% \end{tabular}
% \label{table: boxes_bVars}
% \end{table}

% \begin{itemize}
%     \item Monolithic TR faster than Partitioned TR.
%     \item preprocessing Graph of Utility to compute reachable state and modifying the regret synthesis algorithm to only reason over state reachable form the initial state gives atleast an order of magnitude speed up.
%     \item talk about boolean variables blow-up with increase in \# of objects as Shown in Table \ref{table: boxes_bVars}
% \end{itemize}

\section{Conclusion}
\label{sec: conclusion}

% Symbolic graphs and algorithms are powerful tools for reasoning over systems with large state spaces.
In this work, we provide fundamental algorithms for symbolic value iteration and symbolic regret-minimizing strategy synthesis using ADDs.
% Further, we compare two of the most commonly used Transition Relation representations. 
We benchmark our implementation
% the efficacy of our framework
for robotic manipulation scenarios and show that we are able to achieve significant speedups. In future work, we plan to explore non-zero-sum games and alternate notions of strategies to relax the adversarial assumption and model a strategic human with their own objective.

% We discuss our findings and summarize which Transition relation representation is more suitable for robotic applications.
% \vspace{-1mm}

\bibliographystyle{IEEEtran}
\bibliography{references}

\newpage
\end{document}